\documentclass[journal]{IEEEtran}

\usepackage[noadjust]{cite}
\usepackage{graphicx}
\usepackage{amsmath}
\usepackage{amssymb}
\usepackage{multirow}
\usepackage{array}
\usepackage[letterpaper=true,colorlinks,bookmarks=false]{hyperref}
\usepackage{subfigure}
\usepackage[lined,boxed, ruled]{algorithm2e}
\usepackage{bigstrut}
\usepackage{rotating}

\begin{document}

\title{Person Re-Identification by Semantic Region Representation and Topology Constraint}

\author{Jianjun~Lei,~\IEEEmembership{Senior Member,~IEEE,}
		 Lijie~Niu,
		 Huazhu~Fu,~\IEEEmembership{Senior Member,~IEEE,}
		 Bo~Peng,

		 Qingming~Huang,~\IEEEmembership{Fellow,~IEEE,}
		 and~Chunping~Hou
\thanks{This work was supported in part by the National Natural Science Foundation of China (No. 61722112, 61520106002, 61332016, 61620106009, 61731003), and National Key R\&D Program of China (No. 2017YFB1002900). Copyright~\copyright~2018 IEEE. Personal use of this material is permitted. However, permission to use this material for any other purposes must be obtained from the IEEE by sending an email to pubs-permissions@ieee.org. (Corresponding author: H.~Fu)}
\thanks{J.~Lei, L.~Niu, B.~Peng and C.~Hou are with the School of Electrical and Information Engineering, Tianjin University, Tianjin 300072, China (e-mail: jjlei@tju.edu.cn).}
\thanks{H.~Fu is with the Institute for Infocomm Research, Agency for Science, Technology and Research, Singapore 138632. (e-mail: huazhufu@gmail.com).}
\thanks{Q.~Huang is with the School of Computer and Control Engineering, University of Chinese Academy of Sciences, Beijing 100190, China (e-mail: qmhuang@ucas.ac.cn).}
\thanks{Digital Object Identifier}

}

\markboth{IEEE TRANSACTIONS ON CIRCUITS AND SYSTEMS FOR VIDEO TECHNOLOGY}%
{Shell \MakeLowercase{\textit{et al.}}: Bare Demo of IEEEtran.cls for IEEE Journals}

\maketitle


\begin{abstract}
Person re-identification is a popular research topic which aims at matching the specific person in a multi-camera network automatically. Feature representation and metric learning are two important issues for person re-identification. In this paper, we propose a novel person re-identification method, which consists of a reliable representation called Semantic Region Representation (SRR), and an effective metric learning with Mapping Space Topology Constraint (MSTC). The SRR integrates semantic representations to achieve effective similarity comparison between the corresponding regions via parsing the body into multiple parts, which focuses on the foreground context against the background interference. To learn a discriminant metric, the MSTC is proposed to take into account the topological relationship among all samples in the feature space. It considers two-fold constraints: the distribution of positive pairs should be more compact than the average distribution of negative pairs with regard to the same probe, while the average distance between different classes should be larger than that between same classes. These two aspects cooperate to maintain the compactness of the intra-class as well as the sparsity of the inter-class. Extensive experiments conducted on five challenging person re-identification datasets, VIPeR, SYSU-sReID, QUML GRID, CUHK03, and Market-1501, show that the proposed method achieves competitive performance with the state-of-the-art approaches.
\end{abstract}

\begin{IEEEkeywords}
Person re-identification, part-based feature, metric learning, topological relationship
\end{IEEEkeywords}

\IEEEpeerreviewmaketitle

\section{Introduction}

\IEEEPARstart{T}{he} recent advancement of imaging sensor technology has remarkably increased the adoption of video analytic systems for various applications ranging from home to border surveillance~\cite{Garcia2017tip,Pang2016TIE}. However, due to privacy and maintenance cost concerns, most cameras in such surveillance networks are widely spaced so that their fields-of-views are non-overlapping, and the surveillance system cannot keep consistently tracking in these blind areas~\cite{Wang2016TCSVT,Zhang2016arxiv}. Thus, automatically associating pedestrians in such networks is a key technology for tracking~\cite{Chen2016TCSVT,An2017TCSVT}, which is known as the person re-identification problem. It is of increasing importance in visual surveillance in recent years~\cite{Zhu2017TCSVT,Zheng2016IJCV}. The goal of re-identification~\cite{Zhu2017TIP} is to find person who appears at different location and time in the camera network, in which feature representation and distance metric learning are two fundamental elements~\cite{Liu2015Neuro,Tao2017TCSVT,Zheng2014TIP}.

In the feature extraction stage, due to the various changes in pose, illumination, and viewpoint~\cite{Jing2017TIP,Liu2013ICIP,Zheng2018TPAMI}, the extracted feature should be discriminative and robust to distinguish pedestrians in different views. Some general features~\cite{Yang2014ECCV,Prates2016ICIP,Matsukawa2016CVPR} are proposed to advance the person re-identification research. However, these features extracted by the general procedures are not reliable due to the neglect of misalignment, which leads to a serious mismatching problem. To solve this problem, Lin \emph{et al.}~\cite{Lin2017TIP} encoded the spatial correspondence pattern constrained by a camera pair to guide the patch matching, which requires a number of sample pairs to learn the relationship. There may be some exceptions that do not satisfy the relationship in the uncontrolled situation. Liao \emph{et al.}~\cite{Liao2015CVPR} partitioned the image into several horizontal stripes and analyzed the horizontal occurrence of local features to solve the misalignment problem. Different from these methods, we are more concerned with utilizing the semantic information of the image to tackle the misalignment problem.

For the metric learning, it is a matching rule that yields higher matching score for the image pairs belonging to the same person than the pairs belonging to different pedestrians. Since the high-dimension visual features contain lots of invalid information, the metric should pay more attention to the reliable part in the large scale dimensional features. The success of the system depends on effective distance metric to a large extent. The general procedure of metric learning based methods aims to make the samples of same class closer and the samples of different classes more separable. Triplet constraint~\cite{Ye2016TMM,Chen2016CVPR} was utilized to restrict the distances between negative pairs to be larger than those between positive pairs for the same probe image. Besides, some methods~\cite{Tao2016TIP,Li2016TCSVT} maximized the ratio of inter-class variation and intra-class variation to make the distribution more sparse. In our work, the topological relationship among all samples in the feature space is taken into account to enhance the discrimination between probe samples and gallery images.

In this paper, we propose a person re-identification method based on semantic region representation (SRR) and Mapping Space Topology Constraint (MSTC). The contributions of the proposed method to the person re-identification problem are summarized as follows. To address the misalignment problem, the body of the pedestrian is parsed into semantic parts, which avoids the interference of background and achieves the similarity comparison between corresponding regions. The global and local features are extracted to the SRR to derive more comprehensive and reliable representation of the pedestrian. Besides, a mapping space topology constraint is introduced into metric learning, which exploits the topological relationship among probe and gallery samples to maintain the compactness of intra-class and sparsity of inter-class. In-depth experiments are operated to analyze various aspects of the proposed method, and experimental results show that our method yields competitive performance with the state-of-the-art methods over five challenging benchmarks.

The rest of the paper is organized as follows. Section II gives a brief overview of related works. Section III presents the proposed person re-identification method in detail. We evaluate the performance of the proposed method in Section IV. Finally, the paper is concluded in Section V.

\section{Related works}
Person re-identification methods are generally categorized into two kinds. One focuses on developing efficient and effective feature representation, and the other pays attention to design a proper metric to measure the appearance similarity between images captured in different views.
\subsection{Feature representation}
Feature extraction is a fundamental component in person re-identification. The goal is to design an invariable and stable representation. The strategies can be grouped into two categories, namely, hand-crafted features and deep features. Farenzena \emph{et al.}~\cite{Farenzena2010CVPR} introduced the maximally stable color regions (MSCR), weighted color histograms, and recurrent high-structured pathes (RHSP) to describe the appearance of the pedestrians from different aspects. The three types of features are robust and stable to the changes of views and illumination. Mignon \emph{et al.}~\cite{Mignon2012CVPR} used the 16 bins color histograms in three  color spaces as well as LBP histogram as the descriptors. Chen \emph{et al.}~\cite{Chen2016CVPR} proposed a polynomial feature map which utilizes multiple visual cues for sub-regions. Different feature maps preserve information in different aspects. Su \emph{et al.}~\cite{Su2017TPAMI} integrated the mid-level attributes with low-level feature to describe the images. Besides, the correlative relationship between attributes were utilized to map the original attributes into a continuous space, which improves the accuracy of the representation. Zhao \emph{et al.}~\cite{Zhao2017TPAMI} introduced the saliency information into the matching process which is discriminative and reliable in different views. The corresponding patches with inconsistent saliency information mean the pair of images having different identities.

Given the success of deep learning, high dimensional and informative deep features have gained widely attention. Cheng \emph{et al.}~\cite{Cheng2016CVPR} proposed a novel CNN model which integrates multiple channels to learn the global and local features simultaneously. Zhu \emph{et al.}~\cite{Zhu2017TIP} inserted a hash layer into the network as an intermediate layer, which saves lots of memory and improves retrieval speed. Several subnetwork branches are trained for different horizontal stripes to discriminate pedestrians. Su \emph{et al.}~\cite{Su2016ECCV} trained a deep convolutional neural network on an independent dataset labeled with attributes, and fine-tuned it on a person identification labeled dataset. The learned deep attributes show a well generalization ability in different person re-identification datasets. Wu \emph{et al.}~\cite{Wu2016WACV} concatenated the FC feature with the hand-crafted feature to enhance the effectiveness of FC feature. In~\cite{LiD2017ICCV}, a multi-scale context aware network was proposed to derive a discriminative feature for person re-identification. Zhao \emph{et al}.~\cite{ZhaoH2017ICCV} leveraged the body structure information to guide multi-stage feature decomposition and fusion. In~\cite{Zhao2017ICCV}, the human part-aligned descriptor was provided by utilizing attention models. Su \emph{et al}.~\cite{Su2017ICCV} learned robust descriptors from both global and local parts to alleviate the pose variations. Bak \emph{et al}.~\cite{Bak2017TCSVT} found the corresponding pathes based on appearance similarity, and used a convolutional neural network to extract feature in these patches. Zheng \emph{et al}. ~\cite{Zheng2017Arxiv} aligned pedestrians to standard pose and took the fully connected layer of the fusion network as the descriptor of pedestrian. Sun \emph{et al}.~\cite{Sun2017Arxiv} proposed a refinement strategy to enhance the content consistency within each stripe. However, the long training time and heavy parameter size limit the applications of deep-based methods on mobile and embedded systems.

\begin{figure*}[!t]
\centering
\includegraphics[width=1\linewidth]{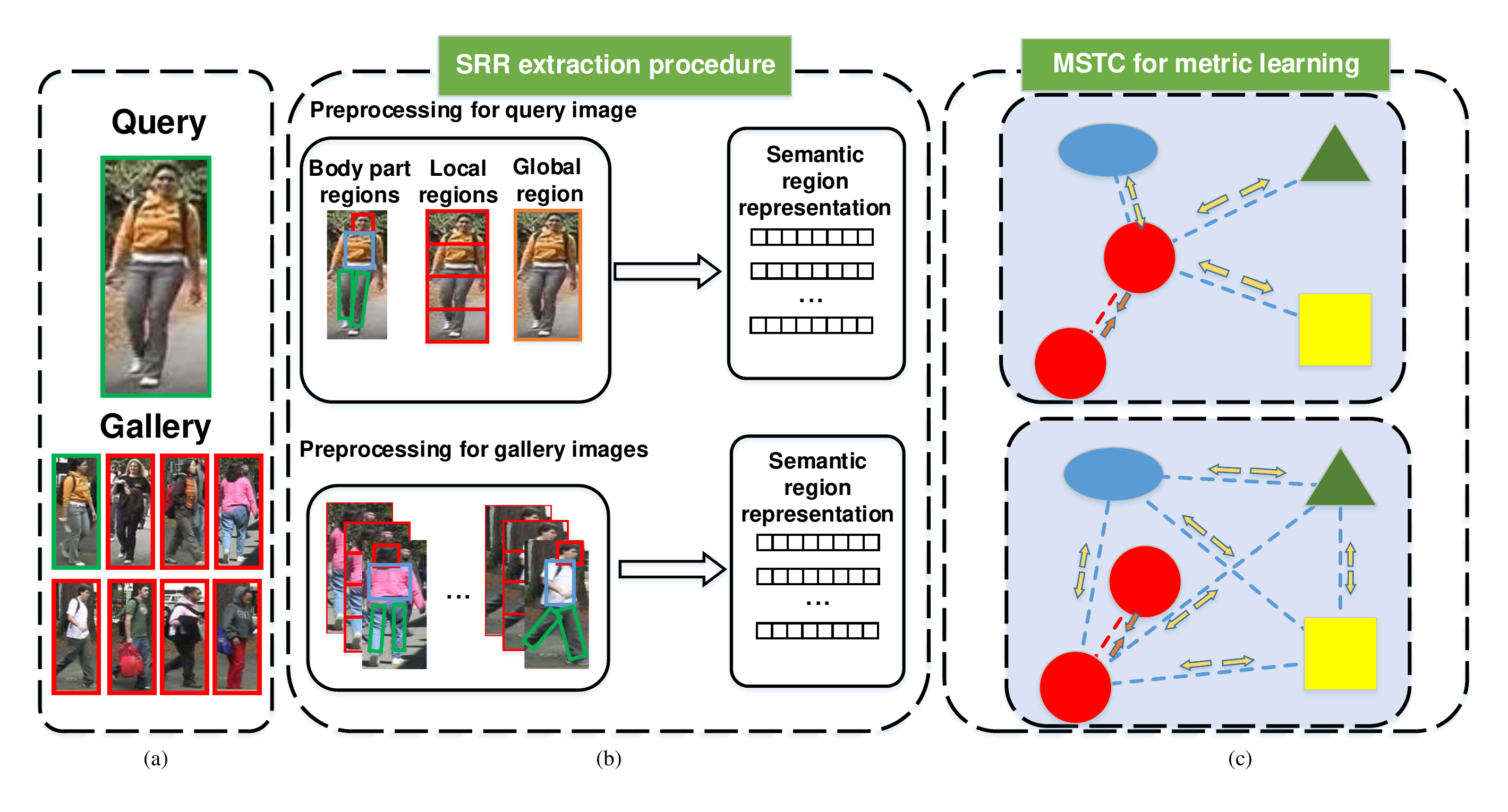}
\caption{The framework of the proposed method. (a) Query is a single-view image, and gallery contains the images captured from the other views. The pictures with the green bounding box are the same identity. The images with red bounding box represents the different identities. (b) The procedure of semantic region representation extraction. Color and texture descriptors are extracted from the body part, local, and global regions. Then PCA operation is used to reduce the dimension of feature vector. (c) The topological relationship constraint for metric learning. The red circle in the center is the probe sample. The same shape represents the matched sample, while different shapes indicate different identities. The upper part shows the first ingredient of the constraint, and it forces to shorten the length of red-dashed line while lengthen the length of blue-dashed lines. The lower part shows the second ingredient of the constraint, and it narrows the distance between positive pair and makes different classes more dispersed.}
\label{frame}
\end{figure*}

\subsection{Metric learning}
Besides robust features, metric learning is the other important part in the re-identification system. The general idea of metric learning is to pull the samples of the same class closer, while push the samples of different classes apart. Euclidean distance is a common method. While it equally treats each dimension information in high dimensional features, which cannot highlight useful information. Thus, most metric learning methods focus on Mahalanobis distance. It transfers the features into a discriminative feature space by utilizing a linear transformation matrix, then computes the Euclidean distance in the discriminative subspace~\cite{Wang2014TCSVT}. KISSME algorithm~\cite{Hirzer2012CVPR} is a popular metric learning method which is based on the Mahalanobis distance. It enlarges the difference between intra-class and inter-class by maximizing the likelihood ratio. Zhang \emph{et al.}~\cite{ZhangL2016CVPR} used a null Foley-Sammon transform to derive a space where the distances between the positive pairs are zero while the distances between the negative pairs are positive. Li \emph{et al.}~\cite{Li2016ICIP} used the AdaBoost algorithm to assemble a set of weak classifiers into a stronger classifier, and achieved a better performance. Weinberger \emph{et al.}~\cite{Weinberger2005NIPS} proposed a large margin nearest neighbor learning method which sets a perimeter for the positive pairs, and any negative samples within the perimeter will be punished. Martinel \emph{et al.}~\cite{Martinel2015TPAMI} used the dynamic time warping idea to model the transformation of the features belonging to cross-view images, and utilized a random forest classifier to choose feasible warp function component. Liao \emph{et al.}~\cite{Liao2015CVPR} treated the intra-class and inter-class differently and maximized the likelihood ratio to derive the maximal difference. Wang \emph{et al.}~\cite{Wang2016ICIP} argued that if the difference between intra-class and inter-class is large, then the distances between negative sample and a pair of positive samples should be approximate. Therefore, they took both individual similarity constraints and contextual similarity constraints into account to learn the metric. Chen \emph{et al.}~\cite{Chen2016CVPR} applied the triplet loss to reduce the intra-class distance while enlarge the inter-class distance. In~\cite{Zhou2017CVPR}, the pairwise loss and symmetric triplet loss were combined to maximize the inter-class variations and minimize the intra-class variations jointly. Chen \emph{et al}.~\cite{Chen2017ICCV} utilized the log-logistic loss to pull positive pairs close and push negative pairs apart. Bak \emph{et al}.~\cite{Bak2017CVPR} trained the CNN using only intensity images, then learned a color metric to account for the different color distributions in camera pairs. Wen \emph{et al}.~\cite{Wen2016ECCV} combined the softmax loss and center loss to train the CNN. The former forces the deep features of inter-class apart, and the latter pulls the deep features of the intra-class to their center. In~\cite{Jin2017IJCB}, the center loss was implemented to person re-identification, which also combined with identification loss and verification loss. Liu \emph{et al}.~\cite{Liu2017CVPR} utilized the angluar softmax to impose constrain on a hypersphere manifold, which was used in face verification. In \cite{Zhong2017CVPR}, a re-ranking strategy was proposed to further improve the accuracy based on initial ranking results. Hermans \emph{et al}.~\cite{Hermans2017arxiv} combined the triplet loss with hard-sample mining to perform deep metric learning. In this paper, we provide a more comprehensive constraint which maintains the compactness of intra-class and the sparsity of inter-class.

\section{Proposed method}
Fig.~\ref{frame} illustrates the framework of the proposed method. First, each image is preprocessed for deriving the semantic region representation. The pedestrian is decomposed into several body parts, such as head, torso, and legs. In addition, the whole image and horizontal windows are also taken into consideration. Color histogram and texture descriptor are extracted from these areas to make up the semantic region representation. After that, a mapping space topology constraint is proposed to learn the metric, which measures the similarity among images captured in different views. The novel constraint not only forces the distances between negative pairs to be larger than those between positive pairs for the same probe image, but also reduces the intra-class variations and enlarges the inter-class variations. The details are discussed in the following subsections.
\subsection{Semantic Region Representation}
\begin{figure}[!t]
	\centering
	\includegraphics[width=0.85\linewidth]{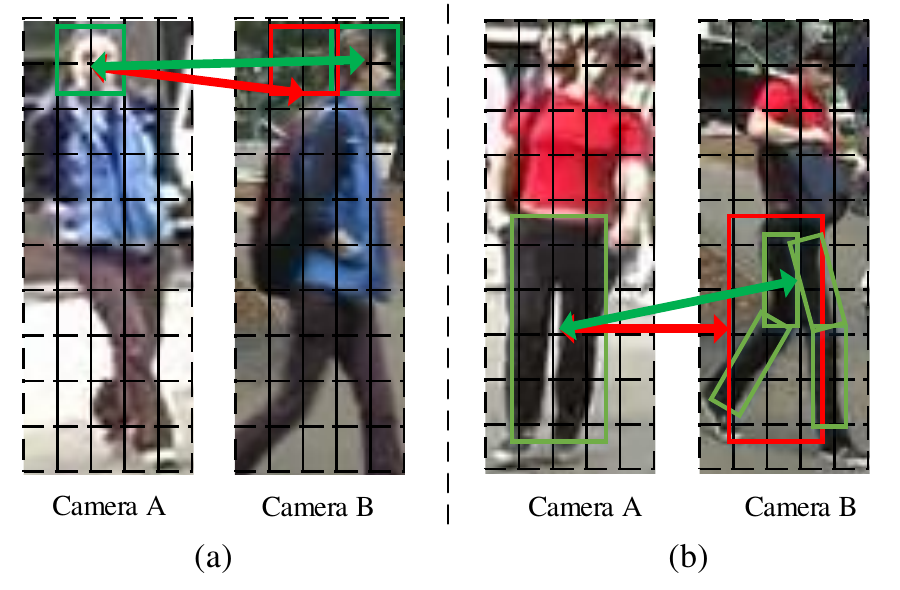}
	\caption{Two examples of spatial misalignment problem, where the images are captured from a pair of camera: A and B. The red boxes in camera B represent the regions which have the same position to the green box in camera A. While the green boxes in camera B indicate the correctly matched areas to the green box in camera A.}
	\label{img-misalign}
\end{figure}

Due to the changes of pose and viewpoint, misalignment is a serious problem for person re-identification. The examples of misalignment are shown in Fig.~\ref{img-misalign}. As shown in Fig.~\ref{img-misalign} (a), the red box in camera B represents the region which has the same position to the green box in camera A. However, the boxes belong to the head and background areas, respectively, and the correct matched region has shifted. The regions which have same position in different views are not corresponding to each other. As shown in Fig.~\ref{img-misalign} (b), the green box in camera A represents the leg of pedestrian, while the region which has the same position in camera B consists of lots of background. There are lots of background interferences in the corresponding regions. To address those issues, the current works~\cite{Liao2015CVPR,Cheng2016CVPR} partitioned the image into several horizontal stripes. While there is serious misalignment in the horizontal direction even in the corresponding horizontal stripe. Some methods tried to solve the misalignment problem by similarity comparison between corresponding patches based on appearance similarity. However, some irrelevant patches may also have the similar appearances which are easily taken as the correspondences.
In view of the factors mentioned above, the semantic region representation (SRR) is proposed to tackle the misalignment problem by utilizing the semantic information. In the SRR, the body part feature avoids the background interference and achieves the accurate similarity comparison between corresponding parts. It pays more attention to the detailed information and is served as a reliable term. The global and local features describe the image from large scale perspective, and maintain the context information. Thus the proposed SRR takes the three types of information into account simultaneously, and the joint representation derives the most comprehensive description.

\begin{figure}[!t]
\centering
\includegraphics[width=0.8\linewidth]{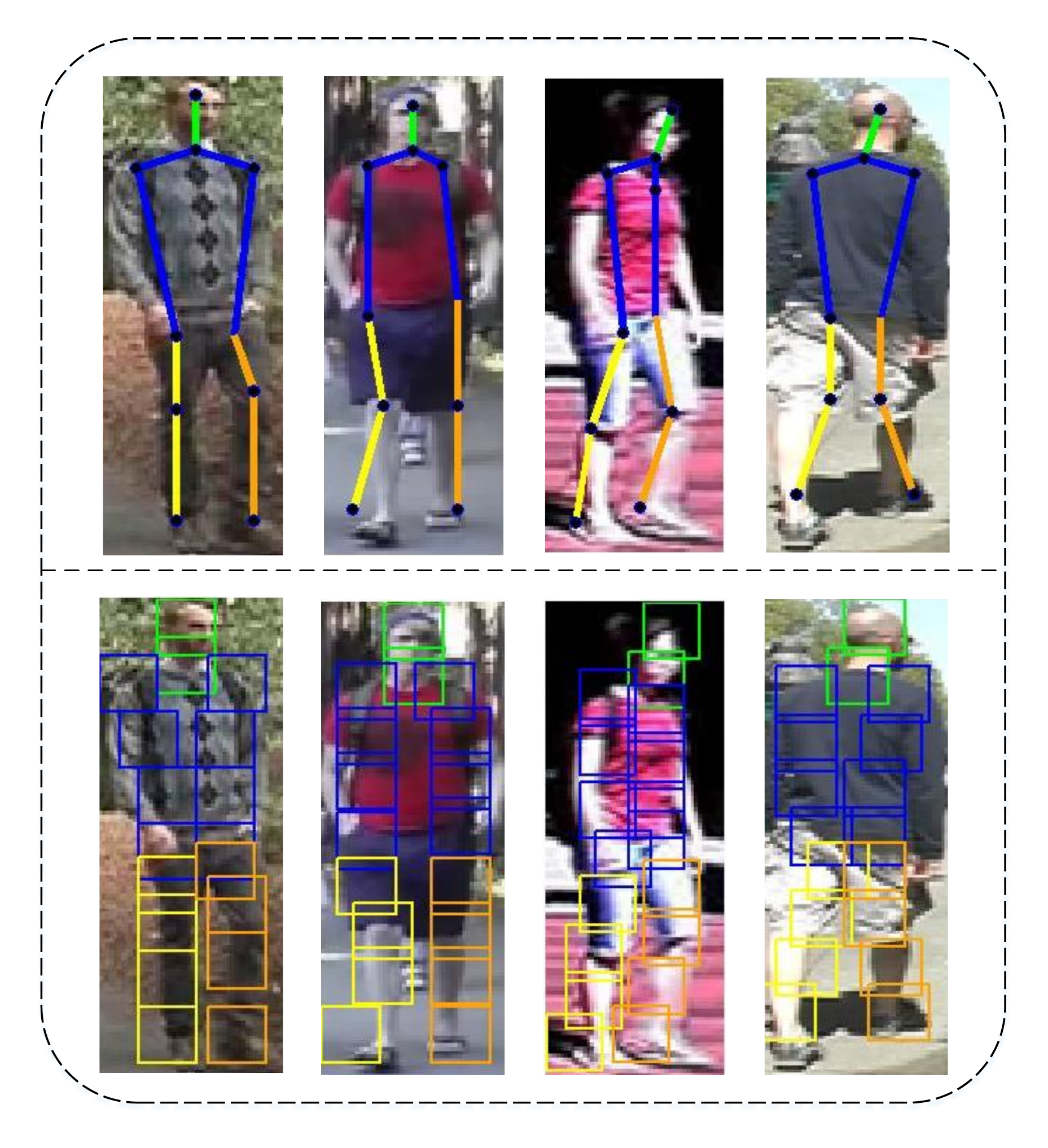}
\caption{Processing images by the implementation of pose estimation. The upper part of this figure shows the detected skeletons of the pedestrians by~\cite{Yang2011CVPR}. The lower part of the figure shows the bounding boxes along the skeleton structure. Different color bounding boxes represent different parts of the body, such as head, torso, and legs.}
\label{img-part}
\end{figure}

In the procedure of feature extraction, each image is processed individually. Considering that the human body is made up of articulated parts rather than a rigid structure, the part-based region feature is more flexible to represent the pedestrian, and the feature extracted in the body part regions derives more reliable information in the foreground against the background interferences. Thus, the pedestrian is parsed into several body parts, such as head, torso, and legs according to~\cite{Yang2011CVPR}, which is trained on Image Parsing Dataset. Each parsed region has explicit semantic information to avoid confusion. The misalignment problem is solved by comparing the similarity of corresponding parts. Fig.~\ref{img-part} shows some processed images. The upper part of the figure shows the detected skeletons of the pedestrians. Along the derived skeleton structure, bounding boxes are assigned to represent the body part regions. The bounding boxes that belong to the same body region have the same color. These colorful bounding boxes locate the head, torso, and legs accurately. Thus, the parsing operation not only avoids the interference of background information, but also alleviates the influence of misalignment. After this operation, a collection of part regions is derived, and the number of the part regions is represented by P. Besides, the image is partition into C horizontal stripes to obtain local region. Since the local regions only consist of the information in one horizontal stripe, the global region is also taken into consideration. Finally, there are (P+C+1) regions for the image, and we use T to represent the number of total regions for convenience.

Within each region, the HSV and LAB color spaces are exploited to extract the dense histogram features, SILTP~\cite{Liao2010CVPR} and HOG are used to capture the texture information. The extracted features that belong to the same region are concatenated together. F is the descriptor of pedestrian, which is equal to ${\rm{F = \{ }}{{\rm{f}}^t}{\rm{\} }}_{t = 1}^{T}$, where the ${\rm{\{ }}{{\rm{f}}^t}{\rm{\} }}_{t = 1}^P$, ${\rm{\{ }}{{\rm{f}}^t}{\rm{\} }}_{t = P+1}^{P+C}$, and ${\rm{\{ }}{{\rm{f}}^t}{\rm{\} }}_{t = P+C+1}$ denote the feature of body part, local and global regions, respectively. Log transform is applied to suppress the large elements of the feature vector, and PCA is used to reduce the dimensionality of the features in each type of regions. The semantic region representation is a combination of these three types of region features. It tackles the background interference and spatial misalignment problem by utilizing the information which is extracted from the semantic regions. Thus, the joint representation gives a good solution to the changes of viewpoints and poses.
\subsection{Mapping Space Topology Constraint}
Person re-identification is treated as an image retrieval problem, and similarity ranking is usually used to find the most relevant pedestrian. The polynomial feature map~\cite{chen2015CVPR} is adopted to compute the similarity score between image descriptors. Since there are several types of regions, the learned metric $W$ is also decomposed into corresponding sub-metrics which are good at comparing the appearance similarity between specific sub-regions. The similarity score between two image descriptors ($F_a$,$F_b$) is the sum of the similarities of different sub-regions,
\begin{equation}\label{eqgai}
\begin{aligned}
  g({F_a},{F_b}) = \sum\limits_{t=1}^{T} { < {\varphi _M}(f_a^t,f_b^t),W_M^t{ > _F} + }  \\
  < {\varphi _B}(f_a^t,f_b^t),W_B^t{ > _F},
  \end{aligned}
\end{equation}
where $<\cdot,\cdot> _F$ is the Frobenius inner product. $< {\varphi _M}(f_a^t,f_b^t),W_M^t{ > _F} = {(f_a^t - f_b^t)^T}W_M^t(f_a^t - f_b^t)$ represents Mahalanobis distance and $< {\varphi _B}(f_a^t,f_b^t),W_B^t{ > _F} = {(f_a^t)^T}W_B^tf_b^t + {(f_b^t)^T}W_B^t{f_a^t}$ corresponds to the unconstrained bilinear form. $W_M^t$ and $W_B^t$ are the matrices that need to be learned in these two distance representations.

In order to achieve the goal of pulling the positive pairs together while pushing the negative pairs away, comprehensive constraints need to be introduced to learn the metric. As mentioned in section II, the metric matrix is seen as a transformation matrix which maps the original feature vector into a discriminative space. The proper distributed samples need to satisfy the basic topological relationship in this space. In this paper, a novel constraint called Mapping Space Topology Constraint (MSTC) is proposed to improve the robustness and discrimination of the metric.

For the sake of convenient expression, the constraint is presented in single-shot case where each pedestrian only has one image in each view, and it can be easily extended to the multi-shot case. The probe set and gallery set are captured in different views, denoted as $\{ X_i\} _{i = 1}^N$ and $\{ Y_i\} _{i = 1}^N$, respectively, where the subscript is the identity of different pedestrians. $N$ indicates the number of identities in the dataset. In order to make the mapping space more discriminative, two ingredients are introduced in the MSTC which emphasize different aspects and reinforce the constraint jointly. For a positive pair ${\rm{\{ }}X_i,Y_i{\rm{\} }}$, the two aspects are as follows.

\begin{equation}\
\left\{ {\begin{array}{*{20}{l}}
{d(X_i,Y_i) \le \frac{{\sum\limits_{j \ne i,\forall j} {d(X_i,Y_j)} }}{N-1}} \\
{d(X_i,Y_i) \le \frac{{\sum\limits_{j \ne k,\forall j,k} {d(Y_j,Y_k)} }}{N(N-1)/2}}
\end{array}} \right. ,
\label{eq-MSTC}
\end{equation}

where $Y_j$ is a sample in gallery set which has different identity with $X_i$. ${d(\cdot,\cdot)}$ refers to the distance between paired samples. Since any different identities in the gallery set constitute a inter-class pair, there are $N(N-1)/2$ pairs in the latter term of Eq.~(\ref{eq-MSTC}).

The first term in Eq.~(\ref{eq-MSTC}) mainly considers the relationship between red and blue-dashed lines in the upper part of Fig.~\ref{frame} (c), which forces the average relative distances between negative pairs be larger than the positive pairs with regard to the same probe. By taking the distribution of the whole training set into account, it not only reduces the distance between positive samples, but also keeps negative samples away from the probe sample. In order to make the sample distribution more discriminative, the second term in Eq.~(\ref{eq-MSTC}) is introduced to enhance the constraint from another perspective. It mainly considers the relationship between red and blue-dashed lines in the lower part of Fig.~\ref{frame} (c), which forces the distance between matched pairs be smaller than the average inter-class distances. Then, the intra-class distance is further reduced and samples of different classes are more dispersed.

Appearance similarity score instead of distance is used to indicate the relationship between samples. The closer the sample pair is, the higher score they will get. Thus, the two relaxed loss function is:
\begin{small}
\begin{equation}
{L_1}(W) \!=\! \frac{1}{N}{\sum\nolimits_{i = 1}^N {[{\alpha _1} \!-\! g(X_i,Y_i) \!+\! \frac{{\sum\limits_{j \ne i,\forall j} {g(X_i,Y_j)} }}{N-1}]} _ + },
\label{eq_tript}
\end{equation}
\end{small}
\begin{small}
\begin{equation}
\begin{aligned}
{L_2}(W) = \frac{1}{N}{\sum\nolimits_{i = 1}^N {\left[ {{\alpha _2} - g(X_i,Y_i) + \frac{{\sum\limits_{j \ne k,\forall j,k} {g(Y_j,Y_k)}  }}{N(N-1)/2}} \right]} _ + }.
\end{aligned}
\label{eq_enhance}
\end{equation}
\end{small}

The $[\cdot]_{+}$ indicates the hinge loss. Eq.~(\ref{eq_tript}) requires the average score of negative pairs be smaller than the matched pair by a margin $\alpha_1$, and Eq.~(\ref{eq_enhance}) enforces the average inter-class score be smaller than the intra-class by $\alpha_2$. Different margins have different effects on the system, which is investigated in the experiments.

In addition to the topological relationship constraint, there is an additional regularization item based on the ${l_{2,1}} - norm$~\cite{Liu2013TPAMI} for the corresponding coefficient matrices, which encourages the rows of the matrix to be zero:
\begin{equation}\
R(W) = {\sum\limits_{t=1}^{T} {\left\| {W_M^t} \right\|} _{2,1}} + {\left\| {W_B^t} \right\|_{2,1}}.
\end{equation}

Besides, since the similarity score should be higher when the pair is similar, ${ < {\varphi _M}(f_a^t,f_b^t),W_M^t{ > _F}}$ corresponds to the negative Mahalanobis distance, and $\{W_M^t\} _{t = 1}^{T}$ is a negative semi-definite matrix set.

Finally, the objective function is:
\begin{equation}
\mathop{\min}_{W}{L_1}(W) + {L_2}(W) + \lambda R(W),
\label{eq_obj}
\end{equation}
\begin{center}
  s.t. $ \{ W_M^t\} _{t = 1}^{T} \in {\mathbb{S}_-}$,
\end{center}
where $\lambda$ is a weight to balance the loss items, and ${\mathbb{S}_-}$ denotes the set of negative semi-definite matrices. To derive a discriminative metric $W$ to satisfy the topology constraint, the alternating direction method of multiplier framework~\cite{Boyd2011} is used to computing for the optimum solution, and the problem Eq.~(\ref{eq_obj}) is reformulated into:
\begin{equation}
\mathop{\min}_{W_1,W_2,W_3,W_4}{L_1}({W_1}) + {L_2}({W_2}) + \lambda R(W_3) + {S}({W_4}),
\end{equation}
\begin{center}
  s.t. $ {W_1} = {W_2} = {W_3} = {W_4}$,
\end{center}
where ${W_1}, {W_2}, {W_3}$ and ${W_4}$ are the variables of different sub-problems. $S({W_4})$ is a constraint item which takes infinity if $\{ W_M^t\} _{t = 1}^{T}$ do not belong to negative semi-definite matrix set, and takes zero otherwise. ${W_1}, {W_2}, {W_3}$ and ${W_4}$ are updated alternately by optimizing one variable whilst fixing the rest. Initially, ${W_1}$ and ${W_2}$ are updated based on their gradients while keep other variables fixed. Then ${W_3}$ is updated by the proximity operator~\cite{Kowalski2009ACHA} by keeping ${W_1}$, ${W_2}$ and ${W_4}$ fixed. Finally, keeping other variables fixed, ${W_4}$ is updated by the Euclidean projection~\cite{Dattorro2006AMathS}. The alternate solving procedure is continuing until a maximum number of iteration is reached.

\section{Experiments}
\subsection{Datasets and Experimental Settings}
In our experiments, five widely used datasets VIPeR~\cite{Gray2008ECCV}, SYSU-sReID~\cite{Guo2016ICPR}, QUML GRID~\cite{Chen2009CVPR}, CUHK03~\cite{Li2014CVPR}, and Market-1501~\cite{Zheng2016ICCV} are used to evaluate and compare the performance of the proposed algorithm and multiple state-of-the-arts. All the datasets are obtained in the public places and provide many challenges faced in the real surveillance scene, such as viewpoint, pose, occlusion, background clutter, and illumination changes.

\begin{figure}[!t]
\centering
\includegraphics[width=1\linewidth]{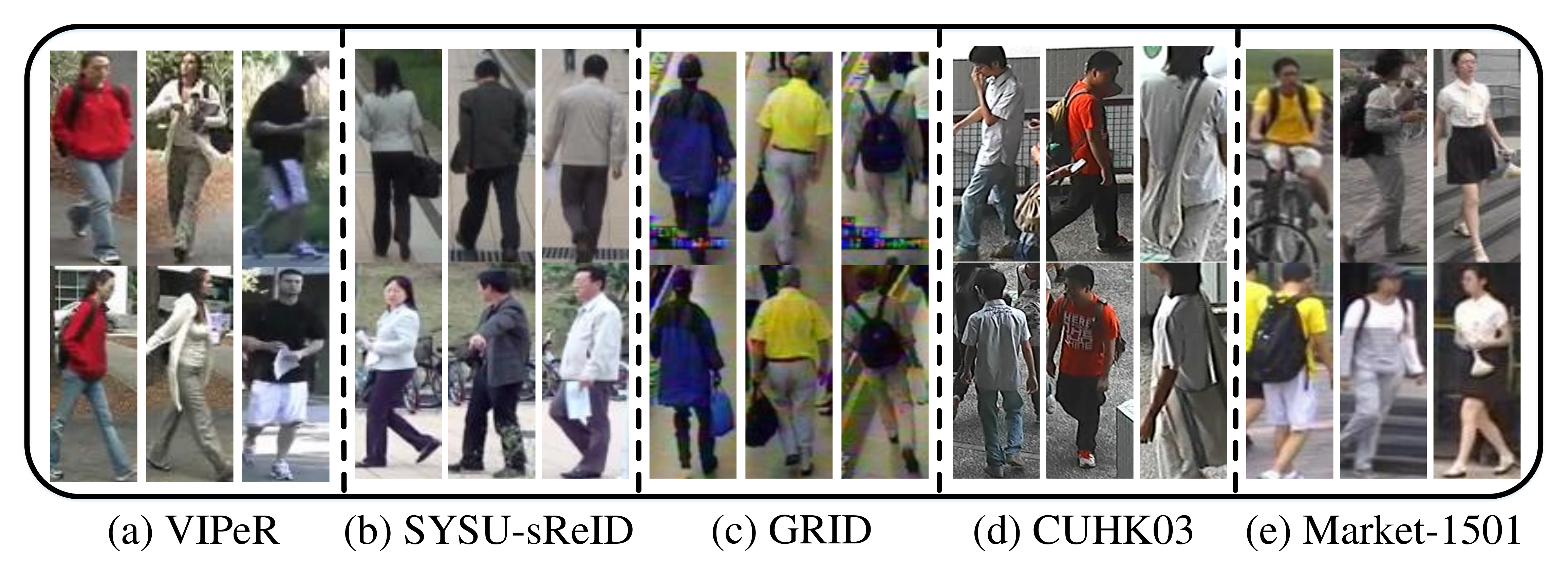}
\caption{Sample images from (a) VIPeR, (b) SYSU-sReID, (c) GRID, (d) CUHK03, and (e) Market-1501. Each column represents the same person from different views.}
\label{exp-dataset}
\end{figure}

The VIPeR dataset~\cite{Gray2008ECCV} is the most widely used dataset for person re-identification. Although lots of efforts have been made, the results on this dataset are still unsatisfactory due to large variations in viewpoints, pose, and illumination. It contains two fixed cameras, each of which captures one image per person. There are 632 image pairs for 632 pedestrians. All the images are scaled to $128 \times 48$ pixels for experiments. Fig.~\ref{exp-dataset}~(a) shows some example pairs of images from the VIPeR dataset. The 632 pairs of images are randomly divided into two half subsets, one half for training and the other for testing. Then the procedure is repeated 10 times, and the average CMC scores are the final result.

The SYSU-sReID dataset~\cite{Guo2016ICPR} is designed for person re-identification specifically. It contains 502 individual pairs taken by two disjoint cameras in a campus environment. Each identity has two images captured in different views. One of the cameras is positioned around a corner, thus there is a wide range of variations, including pose and viewpoint. The identity ambiguity and clutter also introduce interference to the re-identification system. All the images are normalized to $128 \times 48$ pixels in the experiment, and they are partitioned into two half subsets serving as the training set and testing set. 10 random train/test partitions are used for obtaining an average performance. Fig.~\ref{exp-dataset}~(b) shows a snapshot of the sample images in the SYSU-sReID dataset.

The QMUL underGround Re-IDentification (GRID)~\cite{Chen2009CVPR} is collected by 8 disjoint cameras in a busy underground station. There are 250 pedestrian identities, and each identity has two images from different views. Moreover, 775 additional images are used to enlarge the size of gallery set. The additional images do not belong to any samples in the probe set. Illumination, pose, and low image resolution increase the challenges to find the matched target. Fig.~(\ref{exp-dataset})~(c) shows three pairs of images from the datasets.

The CUHK03 dataset~\cite{Li2014CVPR} is one of the largest datasets for person re-identification, which is captured from six disjoint cameras. There are 14097 images of 1467 identities, and each identity is observed by two non-overlapped camera views. The dataset provides two types of images, one is cropped by manual, and the other is detected by pedestrian detector. The experiment is conducted on the latter setting, which is a more realistic setting and could reflect the effectiveness of the algorithm in difficult case. Fig.~(\ref{exp-dataset})~(d) shows some examples of the dataset. The provided partition set in \cite{Li2014CVPR} is used to conduct the experiments, which contains 1367 identities for training and 100 identities for testing, and the procedure is repeated 20 times for the average performance.

The Market-1501 dataset~\cite{Zheng2016ICCV} is collected from six surveillance cameras in Tsinghua University. There are 32668 images of 1501 identities, with each of them captured by two cameras at least and six cameras at most. The provided partition set is used to conduct the experiment, which contains 751 identities for training and 750 identities for testing. There are some example pairs in Fig.~(\ref{exp-dataset})~(e).

The Cumulative Matching Characteristics (CMC)~\cite{Wang2007ICCV} is used to evaluate the performance of the proposed method, which is the most widely used evaluation criteria. The images from the same view are set as the probe set and the other as the gallery set. For each probe sample, it is compared with each image in gallery, and the rank of corrected match is obtained. The expectation of correct match at rank $k$ is represented as rank-$k$ matching rate, and the cumulated values of matching rate at all ranks are the CMC scores. It represents the probability that the matched sample appears in different sized candidate lists. By only recording the first matched target, the evaluation criterion ignores the diversity sizes of datasets.

\begin{figure*}[!t]
\centering
\includegraphics[width=0.8\linewidth]{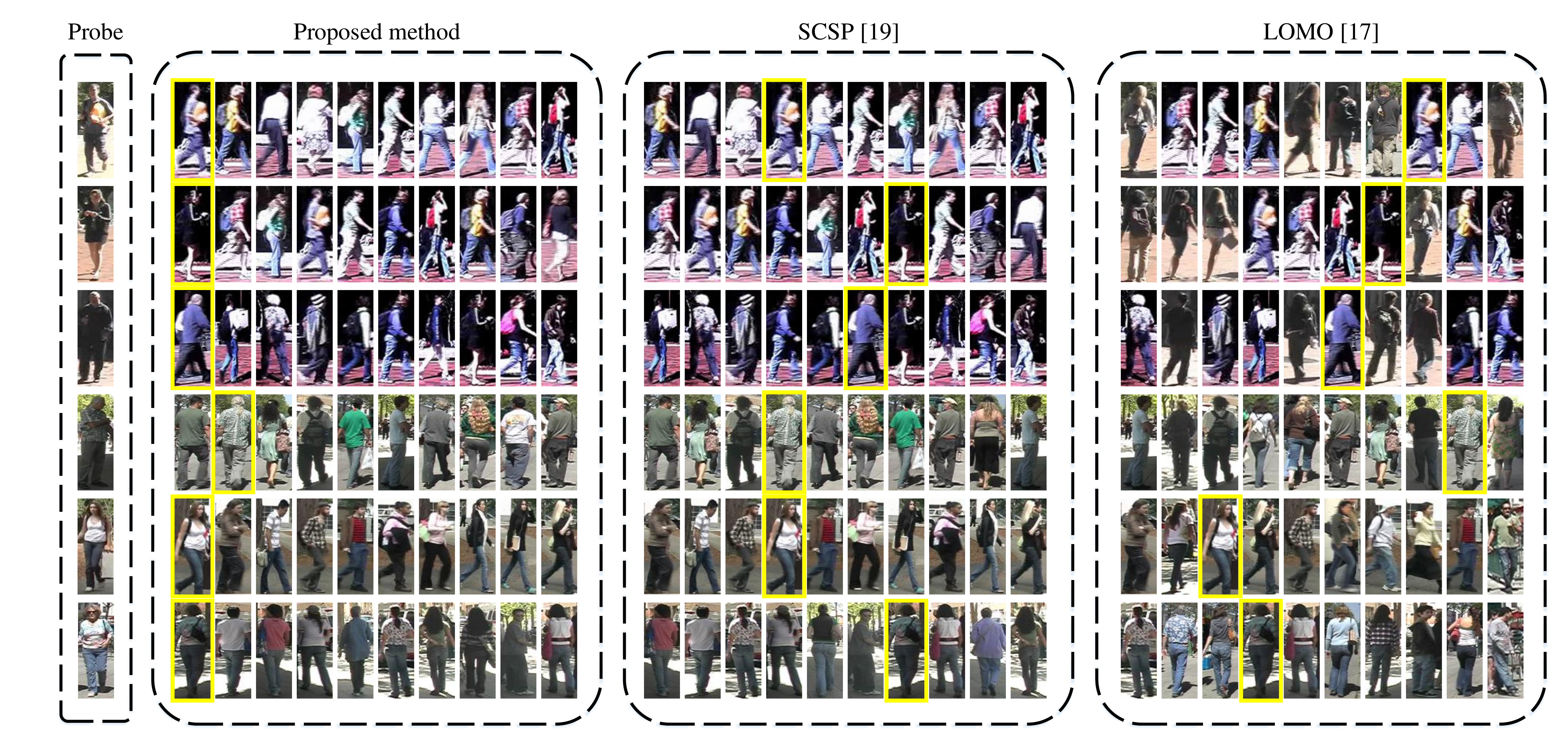}
\caption{Visual results of person re-identification on the VIPeR dataset. The left-most image is the probe sample, and the others are the top ten matched gallery samples sorted by the proposed method, SCSP~\cite{Chen2016CVPR}, and LOMO~\cite{Liao2015CVPR}. The image with yellow box represents the true matched sample. Best viewed in color.}
\label{exp-visual}
\end{figure*}
In the section of feature extraction, due to multiple patterns of occlusion in different views, the regions belonging to arms are easily obscured by the pedestrians themselves. Thus, only the regions of head, torso, left leg, and  right leg are selected, and P is equal to 4 accordingly. As for the local regions, the number of stripes in the image is set as four, i.e., C is equal to 4. The sliding window is used to describe the information in each region. Specifically, the size of the sliding window is $16 \times 8$ pixels, with an overlapping step of 8 pixels in the horizontal direction and 4 pixels in the vertical direction. In each sliding window, color histogram and texture feature are concatenated to preserve a comprehensive description. For color features, two kinds of color histogram are selected. One is 48 bins concatenated histogram and the other is $8 \times 8 \times 8$-bin joint histogram. Each bin represents the occurrence probability of one pattern in the sliding window. The SILTP and HOG serve as the texture descriptors. The dimension reduced by the PCA operation depends on the size of training data, which is set to be 120, 60, 120, 250, 250 for VIPeR, GRID, SYSU-sReID, CUHK03, and Market-1501 dataset respectively. The tradeoff parameter $\lambda$ is set as $3 \times 10^{ - 4}$, which is the same with~\cite{Chen2016CVPR}. Since any two samples with different identities constitute a pair, there is a large number of inter-class pairs in Eq.~(\ref{eq_enhance}). Thus, only 50 images are randomly selected from the training set to reduce the distance of each positive sample pairs.
\subsection{Comparison with State-of-the-Art Methods}

\subsubsection{Result on VIPeR}
Table~\ref{tab-VIPeR} shows the performance comparisons of the proposed method and the state-of-the-art person re-identification methods, including feature learning based methods (i.e. LOMO~\cite{Liao2015CVPR}, SL~\cite{Zhao2017TPAMI}, LCS~\cite{Lin2017TIP}, SLD$^2$L~\cite{Jing2017TIP}, MTL-LORAE~\cite{Su2017TPAMI}), metric learning based methods (i.e. MLCS~\cite{An2017TCSVT}, LSSCDL~\cite{ZhangY2016CVPR}, DR-KISS~\cite{Tao2016TIP}, SCSP~\cite{Chen2016CVPR}, KCVDCA~\cite{ChenYC2016TCSVT}, MARPML~\cite{Li2016TCSVT}, KLFDA~\cite{Xiong2014ECCV}), and deep learning based methods (i.e. DHSL~\cite{Zhu2017TCSVT}, SSDAL~\cite{Su2016ECCV}, MCPCM~\cite{Cheng2016CVPR}, MuDeep~\cite{Qian2017ICCV}, JSLAM~\cite{Peng2018TPAMI}, DLPA~\cite{Zhao2017ICCV}, PDC~\cite{Su2017ICCV}, LDCA~\cite{LiD2017ICCV}, Spindle~\cite{ZhaoH2017ICCV}, IEIT~\cite{Jin2017IJCB}, DPML~\cite{Bak2017TCSVT}). The rank-1, 5, 10, and 20 accuracies of these compared algorithms are listed. For the sake of fair comparisons, the experimental results of the compared methods are directly cited from the corresponding papers due to following the same evaluation protocols. It should be noted that some items are not filled in the table because these evaluation results of the methods have not been reported on the corresponding dataset.

\begin{table}[!t]
\centering
\renewcommand{\arraystretch}{1.2}

\caption{CMC scores (in percentage) of different methods on the VIPeR dataset. ``$-$" means that no reported results are available. The best performance is highlighted in bold fonts.}
\begin{tabular}{|c||c|c|c|c|}
	\hline
       Methods                      &           r=1             &           r=5             &           r=10            &           r=20            \\  \hline
MLCS~\cite{An2017TCSVT}             &           34.58           &           67.25           &           80.59           &           90.43           \\
SL~\cite{Zhao2017TPAMI}             &           44.56           &           71.98           &           83.25           &           {$-$}           \\
DHSL~\cite{Zhu2017TCSVT}            &           44.87           &           {$-$}           &           86.01           &           93.70           \\
LCS~\cite{Lin2017TIP}               &           {51.40}         &           {82.50}         &           {91.20}         &           93.50           \\
LSSCDL~\cite{ZhangY2016CVPR}        &           42.66           &           {$-$}           &           84.27           &           91.93           \\
SSDAL~\cite{Su2016ECCV}             &           43.50           &           71.80           &           81.50           &           89.00           \\
DR-KISS~\cite{Tao2016TIP}           &           43.10           &           74.40           &           86.80           &           {94.80}         \\
KCVDCA~\cite{ChenYC2016TCSVT}       &           43.29           &           72.66           &           83.51           &           92.18           \\
MARPML~\cite{Li2016TCSVT}           &           16.20           &           34.50           &           52.80           &           72.50           \\
MCPCM~\cite{Cheng2016CVPR}          &           47.80           &           74.70           &           84.80           &           91.10           \\
LOMO~\cite{Liao2015CVPR}            &           40.00           &           68.20           &           80.51           &           91.08           \\
SCSP~\cite{Chen2016CVPR}            &           53.54           &           82.59           &           {91.49}         &           {96.65}         \\
SLD${^2}$L~\cite{Jing2017TIP}       &           16.86           &           41.22           &           58.06           &           79.00           \\
MTL-LORAE~\cite{Su2017TPAMI}        &           42.30           &           72.20           &           81.60           &           89.60           \\
KLFDA~\cite{Xiong2014ECCV}          &           34.27           &           65.82           &           79.94           &           90.92           \\
MuDeep~\cite{Qian2017ICCV}          &           43.03           &           74.36           &           85.76           &           {$-$}           \\
JSLAM~\cite{Peng2018TPAMI}          &           49.80           &           77.60           &           90.10           &           95.90           \\
DLPA~\cite{Zhao2017ICCV}            &           48.70           &           74.70           &           85.10           &           93.00           \\
PDC~\cite{Su2017ICCV}               &           51.27           &           74.05           &           84.18           &           91.46           \\
LDCA~\cite{LiD2017ICCV}             &           38.08           &           64.14           &           73.52           &           82.91           \\
Spindle~\cite{ZhaoH2017ICCV}        &           53.80           &           74.10           &           83.20           &           92.10           \\
IEIT~\cite{Jin2017IJCB}             &           50.40           &           77.60           &           85.80           &           $-$             \\
DPML~\cite{Bak2017TCSVT}&       51.70&           $-$           &           $-$             &          {$-$}             \\
Ours                                &           \textbf{54.97}  &           \textbf{83.45}  &           \textbf{91.77}  &           \textbf{96.7}   \\ \hline
\end{tabular} \\
\label{tab-VIPeR}
\end{table}

As can be seen in Table~\ref{tab-VIPeR}, the proposed method produces the best results at different ranks, which demonstrates the effectiveness of the semantic region feature and topology constraint. In terms of the accuracy at rank 1, the proposed method achieves the 54.97\% accuracy, leading to a 1.5\% performance gain over the second best one. As to the rank 20 accuracy, the proposed method achieves 96.72\% accuracy. It demonstrates that most of the matched targets will appear in the top 20 rankings, which saves lots of time and effort to find the right one. Compared with the saliency based method~\cite{Zhao2017TPAMI} with the rank-1 recognition rate of 44.56\%, the proposed method achieves an improvement over 10\%. The improvement of performance demonstrates that the semantic region feature in the proposed method provides an explicit and effective semantic information. Compared with the metric learning based method~\cite{Chen2016CVPR}, the proposed method achieves a better result at rank-1, and consistently outperforms it at different ranks. Besides, the proposed method achieves consistent superior CMC scores than the deep learning baselines. Compared with JSLAM~\cite{Peng2018TPAMI}, PDC~\cite{Su2017ICCV}, DLPA~\cite{LiD2017ICCV}, Spindle~\cite{ZhaoH2017ICCV}, and IEIT~\cite{Jin2017IJCB}, which additionally involve large scale person re-identification datasets (e.g. CUHK03, Market-1501) together to train the model for VIPeR, the proposed method still obtains the superior performance without any extra data augmentation.

\begin{table}[!t]
\renewcommand{\arraystretch}{1.2}
\centering
\caption{CMC scores (in percentage) of different methods on the SYSU-sReID dataset. ``$-$" means that no reported results are available. The best performance is highlighted in bold fonts.}
\begin{tabular}{|c|cccc|}
	\hline
	           Methods          &           r=1             &           r=5             &           r=10                &              r=20             \\ \hline
LCS~\cite{Lin2017TIP}           &           49.30           &           {77.90}         &           {88.10}             &           {94.70}             \\
SCSP~\cite{Chen2016CVPR}        &           {58.84}         &           {82.83}         &           {90.56}             &           {95.46}             \\
KCVDCA~\cite{ChenYC2016TCSVT}   &           40.84           &           71.35           &           82.19               &           90.56               \\
LOMO\cite{Liao2015CVPR}         &           {49.60}         &           76.25           &           85.06               &           90.68               \\
KLFDA~\cite{Xiong2014ECCV}      &           33.40           &           65.90           &           76.30               &           86.30               \\
Ours                            &           \textbf{61.99}  &           \textbf{85.66}  &           \textbf{92.27}      &           \textbf{96.14}      \\ \hline
\end{tabular}
\label{tab-sysu}
\end{table}
\begin{table}[!t]
\renewcommand{\arraystretch}{1.2}
\centering
\caption{CMC scores (in percentage) of different methods on the GRID dataset. ``$-$" means that no reported results are available. The best performance is highlighted in bold fonts.}
\begin{tabular}{|c|cccc|}
	\hline
	          Methods            &           r=1           &           r=5           &          r=10           &          r=20           \\ \hline
	  DHSL\cite{Zhu2017TCSVT}    &          21.20          &          {$-$}          &         {54.24}         &         {65.84}         \\
	  SCSP\cite{Chen2016CVPR}    &         {24.56}         &         {44.32}         &         {55.20}         &         {65.84}         \\
	LSSCDL \cite{ZhangY2016CVPR} &          22.40          &          {$-$}          &          51.28          &          61.20          \\
	   SSDAL\cite{Su2016ECCV}    &         {22.40}         &          {$-$}          &          48.00          &          58.40          \\
	  DR-KISS\cite{Tao2016TIP}   &          20.60          &          39.30          &          51.40          &          62.60          \\
	LOMO\cite{Liao2015CVPR}      &          18.96          &         {42.16}         &          52.56          &          62.24          \\
	            Ours             &      \textbf{26.56}     &      \textbf{46.32}     &       \textbf{56.16}    &       \textbf{66.80}    \\ \hline
\end{tabular}
\label{tab-grid}
\end{table}
The qualitative evaluation is presented in Fig.~\ref{exp-visual}, which shows the visual comparison results of the proposed method with SCSP~\cite{Chen2016CVPR} and LOMO~\cite{Liao2015CVPR} on the VIPeR dataset. Six different queries and the corresponding top ten matched gallery samples are listed in the figure, and the image with yellow box represents the true matched sample. The SCSP~\cite{Chen2016CVPR} achieves the second best performance on the dataset, and the LOMO~\cite{Liao2015CVPR} is a popular algorithm for person re-identification which is robust to the changes of viewpoints. As can be seen in the figure, the better ranking results prove the effectiveness of the proposed method. For the first probe sample in Fig.~\ref{exp-visual}, the semantic region feature can well describe the torso of the pedestrian with a yellow bag, which is a useful cue to distinguish him from the other persons. Although the correct matched target is not the rank-1 as to the fourth probe sample, the distance between the matched samples and the probe sample is much shorter than the other two methods. It demonstrates that utilizing the MSTC constraint is an effective strategy to reduce the distance between positive pairs with large appearance variations. Since the MSTC constraint is an independent module, it is easily combined with the deep-learned features to accommodate the future developments. As to the fifth probe sample, there exists serious misalignment phenomenon, which leads the two other methods fail to find the correct matching target. In contrast, the SRR can find the corresponding body part by utilizing the semantic information, which helps the matched samples get a higher score. The visual result confirms that the SRR is more reliable and effective than the feature extracted from the manually designed horizontal windows.

\begin{table}[!t]
\renewcommand{\arraystretch}{1.2}
\centering
\caption{CMC scores (in percentage) of different methods on the CUHK03 dataset with detected setting. ``$-$" means that no reported results are available.}
\begin{tabular}{|c|c|cccc|}
	\hline
~	                                     & Methods                           & r=1     & r=5     & r=10    & r=20                 \\ \hline
\multirow{4}*{\rotatebox{0}{Non-deep}}	 & Ours                              & 72.38   & 87.65   & 91.13   & 95.24                \\
~	                                     & OL-MANS\cite{Zhou2017ICCV}        & 62.71   & 87.59   & 93.80   & 97.55                \\
~	                                     & CSBT\cite{Chen2017ICCV}           & 55.50    & 84.30    & {$-$}   & 98.00                 \\
~	                                     & DNS\cite{ZhangL2016CVPR}          & 54.70   & 84.75   & 94.80   & 95.20                \\
\hline
\multirow{6}*{\rotatebox{0}{Deep}}	     & JSLAM\cite{Peng2018TPAMI}         & 64.20    & 89.10    & 93.40    & 96.10                 \\
~                                        & S2S\cite{Zhou2018TMM}             & 63.58   & 89.17   & 93.75   & 98.25                \\
~	                                     & DLPA\cite{Zhao2017ICCV}           & 81.60    & 97.30    & 98.40    & 99.50                 \\
~	                                     & PDC\cite{Su2017ICCV}              & 78.29   & 94.83   & 97.15   & 98.43                \\
~	                                     & LDCA\cite{LiD2017ICCV}            & 67.99   & 91.04   & 95.36   & 97.83                \\
~	                                     & Spindle\cite{ZhaoH2017ICCV}        & 88.50    & 97.80    & 98.60    & 99.20                 \\
 \hline
\end{tabular}
\label{tab-cuhk03}
\end{table}
\begin{table}[!t]
\renewcommand{\arraystretch}{1.2}
\centering
\caption{CMC scores (in percentage) of different methods on the Market-1501 dataset. ``$-$" means that no reported results are available.}
\begin{tabular}{|c|c|cccc|}
	\hline
~                                           &Methods                            & r=1       & r=5     & r=10    & r=20            \\ \hline
\multirow{4}*{\rotatebox{0}{Non-deep}}      & Ours                              & 67.04     & 82.95   & 88.60   & 92.73         \\
~                                           & OL-MANS\cite{Zhou2017ICCV}        & 60.67     & {$-$}   & {$-$}   & 91.87         \\
~                                           & CSBT\cite{Chen2017ICCV}           & 42.90     & {$-$}   & {$-$}   & {$-$}          \\
~                                           & DNS\cite{ZhangL2016CVPR}          & 61.02     & {$-$}   & {$-$}   & {$-$}         \\
\hline
\multirow{8}*{\rotatebox{0}{Deep}}          & JSLAM\cite{Peng2018TPAMI}         & 65.70     & {$-$}   & {$-$}   & {$-$}          \\
~                                           & S2S\cite{Zhou2018TMM}             & 65.32     & {$-$}     & {$-$}     & {$-$}         \\
~                                           & DLPA\cite{Zhao2017ICCV}           & 81.00     & 92.00     & 94.70      & {$-$}          \\
~                                           & PDC\cite{Su2017ICCV}              & 84.14     & 92.73     & 94.92     & 96.82         \\
~                                           & LDCA\cite{LiD2017ICCV}            & 76.25     &{$-$}      & {$-$}     & {$-$}         \\
~                                           & Spindle\cite{ZhaoH2017ICCV}       & 76.90     & 91.50     & 94.60     & 96.70         \\
~                                           & CamStyle\cite{Zhong2018CVPR} & 89.49     &{$-$}      & {$-$}     & {$-$}    \\
~                                           & HA-CNN\cite{Li2018CVPR}           & 91.20     &{$-$}      & {$-$}     & {$-$}           \\
 \hline
\end{tabular}
\label{tab-market1501}
\end{table}
\subsubsection{Result on SYSU-sReID}

\begin{table*}[!t]
\renewcommand{\arraystretch}{1.2}
  \centering
  \caption{The top matching rates (in percentage) of three variants on the VIPeR, SYSU-sReID, GRID, CUHK03, and Market-1501 datasets}
    \begin{tabular}{|c|ccc|ccc|ccc|ccc|ccc|}
    \hline
    Methods & \multicolumn{3}{c|}{VIPeR} & \multicolumn{3}{c|}{SYSU-sReID} & \multicolumn{3}{c|}{GRID} & \multicolumn{3}{c|}{CUHK03} & \multicolumn{3}{c|}{Market-1501} \bigstrut\\
\cline{2-16}          & r=1   & r=5   & r=10  & r=1   & r=5   & r=10  & r=1   & r=5   & r=10 & r=1   & r=5   & r=10 & r=1   & r=5   & r=10 \bigstrut\\
    \hline
    SRR   & 54.68 & \textbf{83.64} & 91.68 & 60.72 & 85.30  & 92.19 & 26.24 & 44.24 & 54.44& 69.66 & 83.26 & 87.66 & 64.96 & 82.48 & 88.06 \bigstrut[t]\\
    MSTC  & 53.70  & 82.78 & 91.36 & 59.20  & 83.78 & 91.24 & 26.14 & 45.12 & 55.60 & 68.41 & 81.59 & 88.07 & 65.11 & 82.42 & 87.95 \\
    SRR\_MSTC & \textbf{54.97} & 83.45 & \textbf{91.77} & \textbf{61.99} & \textbf{85.66} & \textbf{92.77} & \textbf{26.56} & \textbf{46.32} & \textbf{56.16} & \textbf{72.38} & \textbf{87.65} & \textbf{91.13} & \textbf{67.04} & \textbf{82.95} & \textbf{88.60} \bigstrut[b]\\
    \hline
    \end{tabular}%
  \label{tab-comp}%
\end{table*}%
Five approaches are selected to compare with our method in Table~\ref{tab-sysu}, namely, LCS~\cite{Lin2017TIP}, SCSP~\cite{Chen2016CVPR}, KCVDCA~\cite{ChenYC2016TCSVT}, LOMO~\cite{Liao2015CVPR}, KLFDA~\cite{Xiong2014ECCV}. The result shows that the proposed method has better re-identification performance than the state-of-the-art methods. It achieves a rank-1 recognition rate of 61.99\%, which outstands the second best one by 3.15\%. Comparing with the LOMO~\cite{Liao2015CVPR}, the proposed method outperforms it by 12.39\% at rank-1 recognition, which indicates the semantic region feature is more reliable than the feature extracted from the manually designed horizontal windows. This is because the semantic region feature can solve the misalignment problem effectively and avoid the background interferences.
\begin{figure}[!t]
	\centering
	\includegraphics[width=1\linewidth]{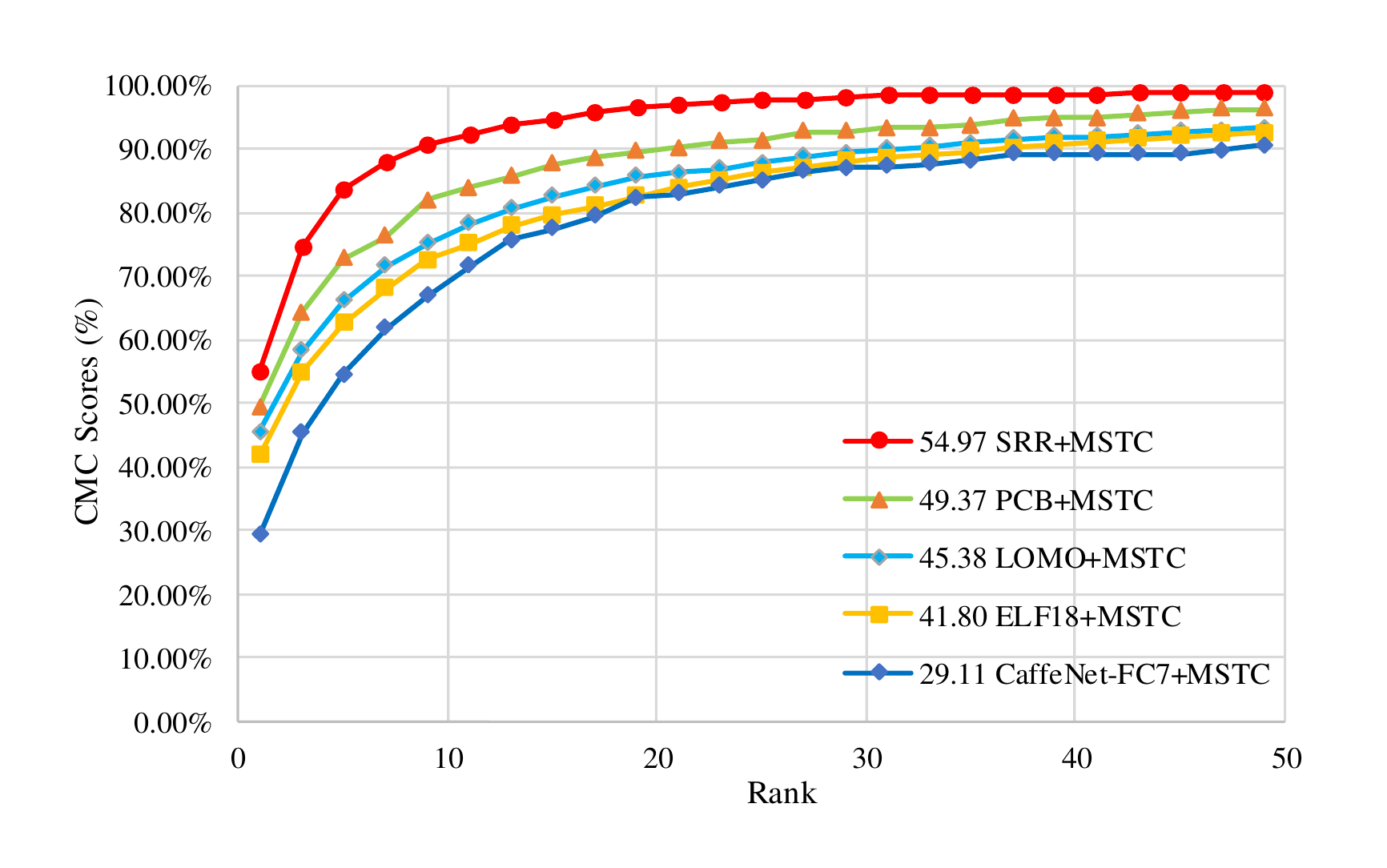}
	\caption{CMC curves and rank-1 identification rates of the proposed SRR and four person re-identification features on the VIPeR dataset.}
	\label{fig-srr}
\end{figure}

\subsubsection{Result on GRID}
The provided training/testing split with 10 trials are used to implement the experiment. 125 image pairs are served as the training set, while the remaining 125 image pairs and the 775 extra images are used for testing in each trial. The proposed method is compared with six state-of-the-art approaches, namely, DHSL~\cite{Zhu2017TCSVT}, SCSP~\cite{Chen2016CVPR}, LSSCDL~\cite{ZhangY2016CVPR}, SSDAL~\cite{Su2016ECCV}, DR-KISS~\cite{Tao2016TIP}, LOMO~\cite{Liao2015CVPR}. Table~\ref{tab-grid} shows the rank-1, 5, 10, 20 matching rate. The proposed method achieves the new state-of-the-art result, with 26.56\% rank-1 accuracy, which is higher than the best result of the other methods by 2\%. Comparing with above datasets, it is easily to find that the task on the GRID is more challenging. It is because the images are captured from 8 cameras, which leads to a large variation on illuminations and viewpoints. Besides, there are 775 additional images which further increase the challenges. However, the proposed method still achieves a promising performance than other methods in the challenging dataset.

\subsubsection{Result on CUHK03}
Due to the misalignment and detector errors, the detected images in CUHK03 dataset could provide more realistic and challenging for the algorithms. Table~\ref{tab-cuhk03} shows the comparison results between the proposed method and nine state-of-the-art person re-identification methods on the CUHK03 dataset with detected setting. The compared methods include non-deep learning baselines (i.e. OL-MANS~\cite{Zhou2017ICCV}, CSBT~\cite{Chen2017ICCV}, DNS~\cite{ZhangL2016CVPR}) and deep learning based methods (i.e. JSLAM~\cite{Peng2018TPAMI}, S2S~\cite{Zhou2018TMM}, DLPA~\cite{Zhao2017ICCV}, PDC~\cite{Su2017ICCV}, LDCA~\cite{LiD2017ICCV}, Spindle~\cite{ZhaoH2017ICCV}). As can be observed from Table~\ref{tab-cuhk03}, the proposed method outperforms the non-deep learning baselines at most ranks, although is slight lower than OL-MANS~\cite{Zhou2017ICCV} and CSBT~\cite{Chen2017ICCV} at rank-20. Compared with the deep learning methods, the proposed method achieves competitive performance, and outperforms some of them at low ranks including JSLAM~\cite{Peng2018TPAMI}, S2S~\cite{Zhou2018TMM}, and LDCA~\cite{LiD2017ICCV}. Although the performance of the proposed method is lower than DLPA~\cite{Zhao2017ICCV}, it still has advantages in fewer parameters. The competitive performance on the large scale dataset confirms the effectiveness of the proposed method, even if there exist distractors due to automatically detected settings.
\begin{figure}[!t]
	\centering
	\includegraphics[width=1\linewidth]{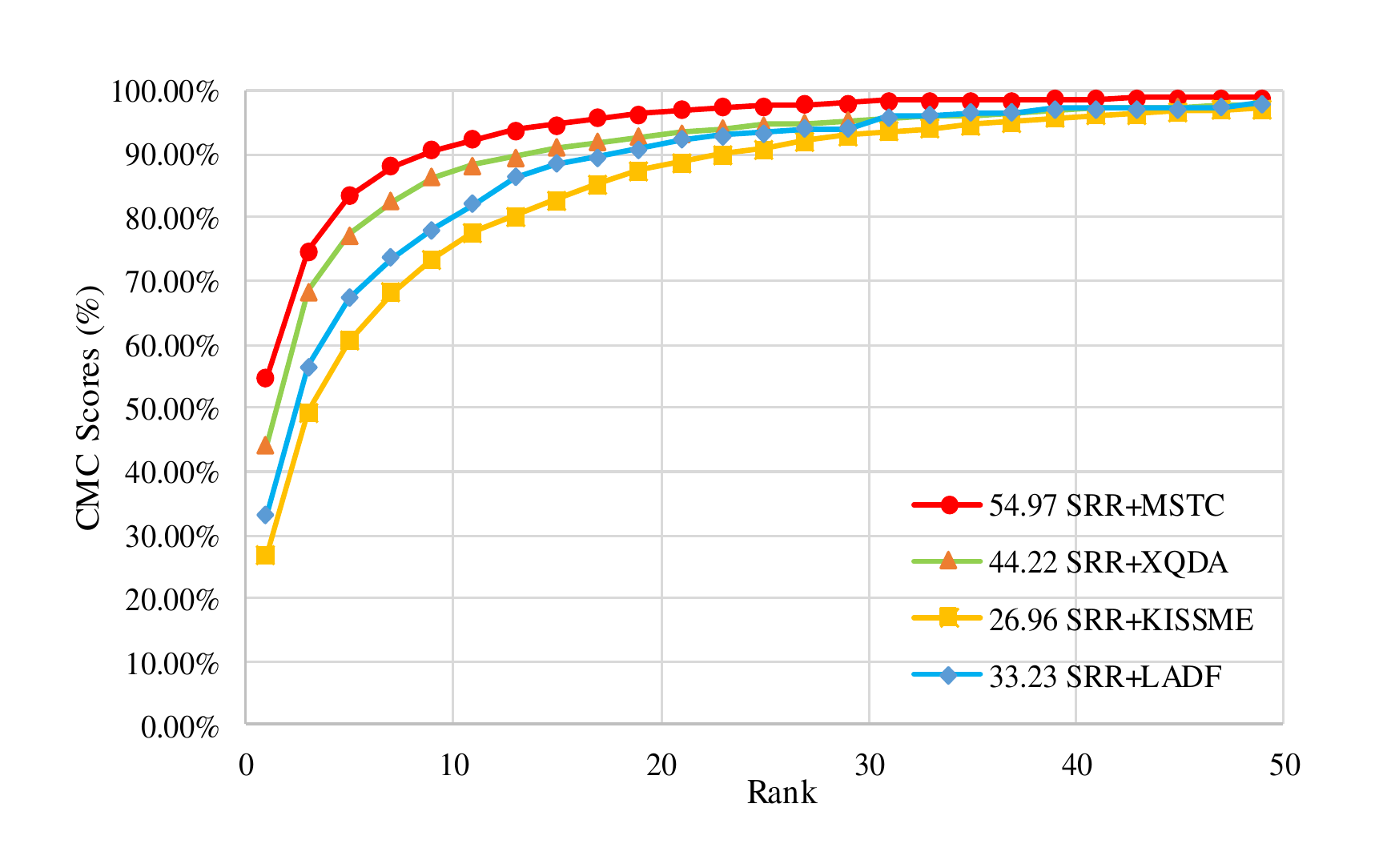}
	\caption{CMC curves and rank-1 identification rates of the proposed MSTC and three metric learning approaches on the VIPeR dataset.}
\label{fig-mstc}
\end{figure}

\subsubsection{Result on Market-1501}
The dataset employs Deformable Part Model to detect pedestrians from surveillance cameras. Besides, there are also some false alarm detection results, which provide a complex environment to evaluate the performance of the algorithms. Table~\ref{tab-market1501} shows the comparison results of the proposed methods and the state-of-the-art approaches on the Market-1501 dataset. As can be observed from the table, the proposed method obtains consistent superior CMC scores than all the non-deep learning baselines (i.e. OL-MANS~\cite{Zhou2017ICCV}, CSBT~\cite{Chen2017ICCV}, DNS~\cite{ZhangL2016CVPR}). Besides, the proposed method also outperforms some deep learning compared methods (i.e. JSLAM~\cite{Peng2018TPAMI} and S2S~\cite{Zhou2018TMM}). Although the performance of the proposed method is lower than some of deep learning based methods such as PDC~\cite{Su2017ICCV}, the proposed method could still be comparable with them with the advantage of fewer quantities of training data and parameter.
\subsection{Ablation study}
\subsubsection{Analysis of Different Components in Proposed Method}
There are two main components in the proposed method, one is the SRR and the other is the MSTC for metric learning. In order to investigate how each component contributes to the person re-identification accuracy, three variants of the proposed method are conducted as follows.

Variant 1 (denoted as SRR): In the feature extraction process, the SRR is used to get a more reliable and comprehensive description. The common triplet constraint $L_1(W)$ in Eq.~(\ref{eq_obj}) is used to learn the metric, which is a simple yet effective constraint to force the distance between positive pair to be smaller than those between negative pairs for the same probe.

Variant 2 (denoted as MSTC): As for the descriptor, the color histogram and texture feature are concatenated to preserve a comprehensive description in four manually designed horizontal stripes and the whole image. The proposed MSTC is used to learn the discriminative metric.

Variant 3 (denoted as SRR\_MSTC): The last variant uses the SRR feature as the image representation, and the MSTC as the constraint to learn the metric. It is the full version of the proposed method, which can be compared with two other variants to investigate the effectiveness.

\begin{table}[!t]
\renewcommand{\arraystretch}{1.2}
\centering
\caption{Rank-1 score (in percentage) of different ingredients in SRR on three datasets. The best performance is highlighted in bold fonts.}
\begin{tabular}{|c|ccc|}
	\hline
Methods                         &           Market-1501             &           CUHK03             &           SYSU-sReID                        \\ \hline
Part                            &           38.87           &           48.48           &           39.04                    \\
Global                          &           57.99           &           62.87           &           46.21                     \\
Local                           &           60.99           &           63.08           &           57.77                   \\
Global+Part                     &           62.03           &           65.39           &           53.39                   \\
Global+Local                    &           65.11           &           68.41           &           59.20                   \\
Local+Part                      &           65.74           &           64.67           &           59.76                     \\
SRR                             &           \textbf{67.04}  &           \textbf{72.38}  &           \textbf{61.99}    \\ \hline
\end{tabular}
\label{tab-srrcomponent}
\end{table}

Experimental results of the three variants on the challenging datasets are shown in Table~\ref{tab-comp}. By applying the variable-controlling approach, it is easy to find the impact of the two components of the proposed method. As can be seen from the table, the SRR\_MSTC shows the most satisfactory performance compared with the other two variants. Comparing the SRR\_MSTC with SRR, it is easy to find that except a slight decline in the rank-5 accuracy on the VIPeR dataset, the MSTC constraint consistently improves the CMC scores in most cases. The results can confirm the effectiveness of the proposed constraint. By comparing SRR\_MSTC with MSTC, it can be found that the SRR feature boosts the accuracy by up to 2\%, which plays an important role in the algorithm. The reason is that the features become more robust and discriminative by utilizing the semantic information.

\subsubsection{Comparison of Feature Extraction Methods}
The proposed SRR are compared with four person re-identification feature extraction methods on the VIPeR dataset, including LOMO~\cite{Liao2015CVPR}, ELF18~\cite{ChenYC2016TCSVT}, CaffeNet-FC7~\cite{Zheng2017TOMM}, and PCB~\cite{Sun2017Arxiv}. LOMO is a high dimensional descriptor which contains color and SILTP histograms. ELF18 consists of RGB, HSV, YCbCr, Lab, YIQ and 16 Gabor texture features, which are extracted from 18 horizontal stripes. CaffeNet-FC7 is extracted from the FC7 layer in the CaffeNet model, which is pre-trained on ImageNet dataset and fine-tuned on the Market-1501. The procedure of fine-tuning is taken as a combination of identification and verification tasks. PCB~\cite{Sun2017Arxiv} is a combination of features which are derived from the stripes regions on the conv-layer. The adopted metric learning is the proposed MSTC, hence, these methods are denoted as SRR+MSTC, LOMO+MSTC, ELF18+MSTC, CaffeNet-FC7+MSTC, and PCB+MSTC, respectively. The person re-identification results are shown in Fig.~\ref{fig-srr}. As shown in the figure, the proposed SRR is much more effective than the compared features, and the CMC scores of the proposed method are higher than the other compared methods up to 6 percentages. It is because that the SRR considers the semantic information, which reduces the background interference and achieves effective similarity comparison between the corresponding regions.

\begin{figure}[!t]
	\begin{center}
		\includegraphics[width=0.73\linewidth]{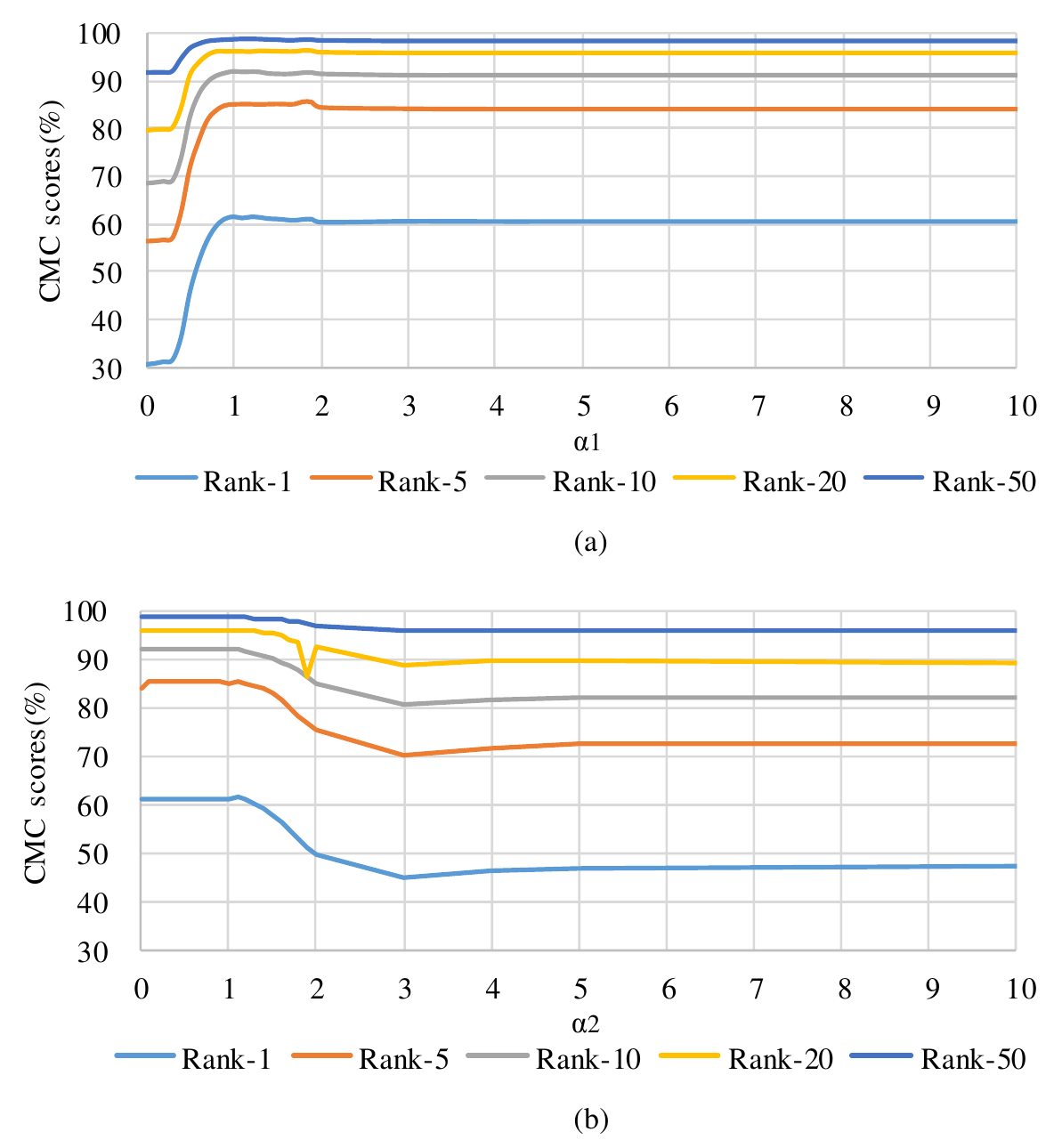}
		\caption{Parameter analysis. CMC scores with respect to (a) parameter $\alpha_1$ when $\alpha_2$ fixed; (b) parameter $\alpha_2$ when $\alpha_1$ fixed. The rank 1,5,10,20,50 accuracies are shown in blue, red, gray, yellow, and purple respectively.}
		\label{img-polar}
	\end{center}
\end{figure}

\subsubsection{Comparison of Metric Learning Methods}
The proposed MSTC is compared with three metric learning approaches on the VIPeR dataset, including XQDA~\cite{Liao2015CVPR}, KISSME~\cite{Hirzer2012CVPR}, and LADF~\cite{LiZ2013CVPR}. The adopted feature uses the proposed SRR, hence, these methods are denoted as SRR+MSTC, SRR+XQDA, SRR+KISSME, and SRR+LADF, respectively. The results of comparison experiments are shown in Fig.~\ref{fig-mstc}. As can be observed from the figure, the proposed MSTC obtains consistent superior CMC scores than the compared alternatives. For example, using MSTC improved the rank-1 recognition rate from 44.22\% by XQDA to 54.97\% on VIPeR. The reason is that the comprehensive MSTC exploits the topological relationship among probe and gallery samples to maintain the compactness of intra-class as well as the sparsity of inter-class.

\subsubsection{Effect of Different Ingredients in SRR}
To explore the effect of different ingredients of the SRR, the experiment is conducted on three datasets. The Rank-1 accuracy is shown in Table~\ref{tab-srrcomponent}. ``Part" means only introducing the semantic parts based features, ``Global+Part'' is the feature extracted from the regions of global and body part, and ``Local+Part'' is the combination of local feature and semantic parts based feature. As can be seen in Table~\ref{tab-srrcomponent}, the CMC scores are further improved when combined with local and global descriptors. This is because semantic part based feature pays more attention to the detailed information, while global and local descriptors describe the pedestrian from the higher scales, which are effective supplements. Thus the proposed SRR takes the three types of information into account simultaneously, and the joint representation derives the most comprehensive description.

\begin{figure}[!t]
	\begin{center}
		\includegraphics[width=0.85\linewidth]{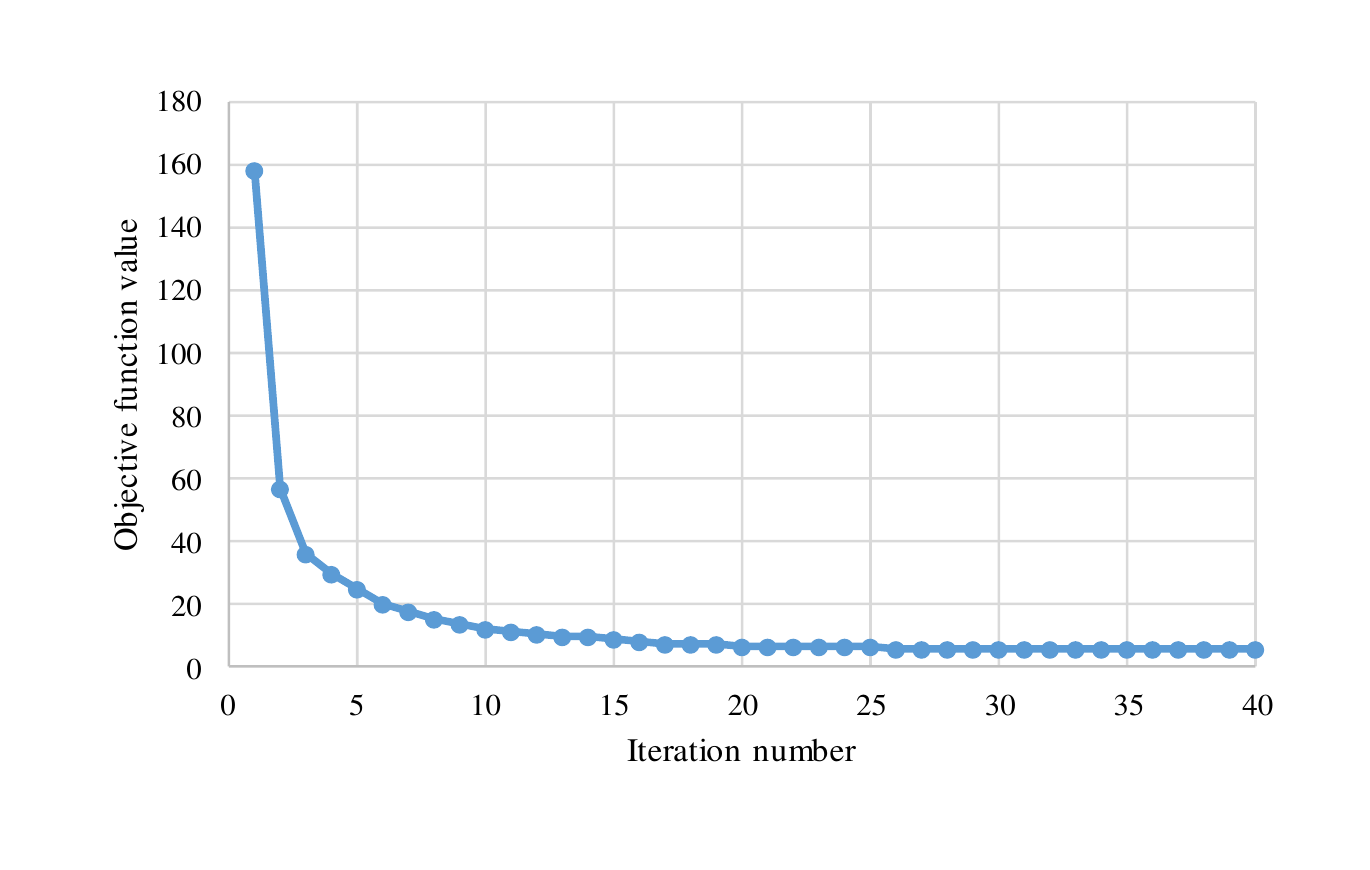}
		\caption{The convergence curve of training process on the SYSU-sReID dataset.}
		\label{img-convergence}
	\end{center}
\end{figure}

\subsection{Discussion}
\subsubsection{Effects of different margin settings}
The experiment is conducted on the SYSU-sReID dataset to investigate the effects of different margins on the rank accuracy. The 502 individual pairs are divided into two half subsets, where 251 pairs of them constitute the training set and the remaining constitute the testing set. The final experimental result is derived by repeating the procedure 10 times. Since the two ingredients in the topology constraint focus on different aspects, they should not be treated equally. The constraint will place more emphasis on one aspect when the margin becomes large. Thus, the margin is used to determine the balance of two items in the constraint instead of using weights. The impact of one item on the performance will be studied by fixing the other one.

The results are shown in Fig.~\ref{img-polar}, where the trends of different rank accuracies are shown in different colorful curves. Fig.~\ref{img-polar} (a) shows the impact of $\alpha_1$ on CMC scores with $\alpha_2$ fixed. The result maintains a promising performance when $\alpha_1$ is large. This is because the larger $\alpha_1$ can enlarge the distances between negative pairs as well as compress those between positive pairs with regard to the same probe. Since the discrimination between different samples will descend with the decreasing of the $\alpha_1$, the accuracy declines when $\alpha_1$ is less than 1. Fig.~\ref{img-polar} (b) shows the result with $\alpha_1$ fixed. As shown in the figure, there will be a great drop in accuracy when $\alpha_2$ increases. The reason is that the larger $\alpha_2$ in the second item will cause over-fitting and decrease the generalization ability. The rank accuracy achieves the best result when $\alpha_2$ equals 1.1. Thus, $\alpha_1$ and $\alpha_2$ are set as 1 and 1.1 in the experiments to get a satisfactory performance.
\begin{figure}[!t]
	\begin{center}
		\includegraphics[width=0.50\linewidth]{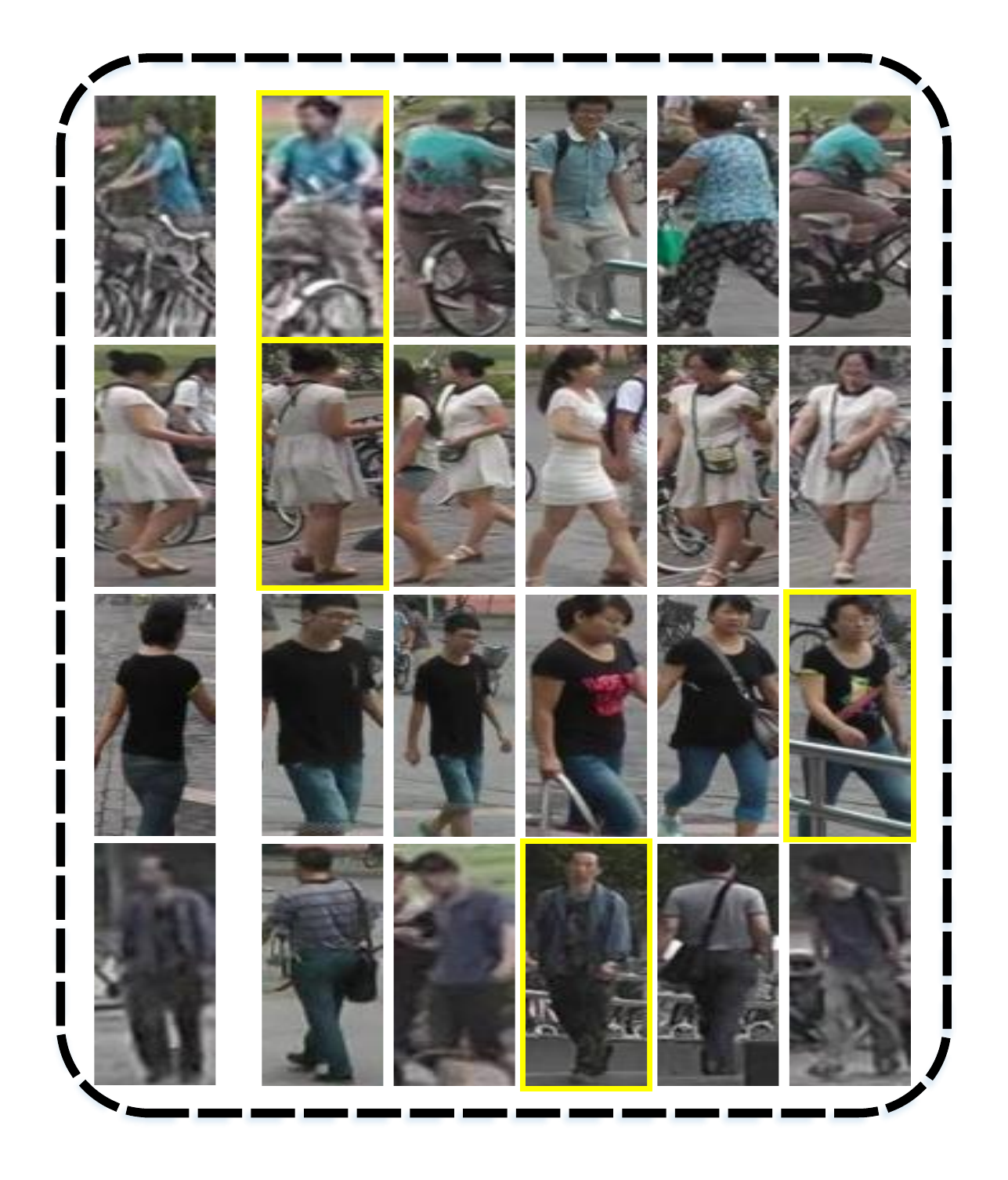}
		\caption{Some challenging examples for our proposed method. The left-most image is the probe sample, and the others are the top five matched gallery samples sorted by the proposed method. The image with yellow box represents the true matched one. Best viewed in color.}
		\label{img-challengecase}
	\end{center}
\end{figure}
\subsubsection{Convergence Analysis}
The objective function in Eq.~\ref{eq_obj} is a convex optimization problem, which is optimized alternatively with respect to one variable while fixing the others. The alternating direction method used in the paper combines the superior convergence properties of the multiplier method with the decomposability of dual ascent~\cite{Boyd2011}. Therefore, the optimal solution can be derived via alternating optimization. To investigate the convergence of the MSTC, the value of the loss function during the training process are shown in Fig.~\ref{img-convergence}. The experiment is conducted on the SYSU-sReID dataset. As can be seen from Fig.~\ref{img-convergence}, the value of loss function drops quickly and reaches the stability after several iterations, which demonstrates the convergence of the proposed method in practice.
\subsubsection{Computational Complexity}
The proposed method is implemented in MATLAB on a Quad core 3.6GHz computer. Taking the VIPeR dataset as an example, the time consumed for feature extraction is 0.3 seconds per image. As a comparison, LOMO~\cite{Liao2015CVPR} takes $0.02s$, ELF18~\cite{ChenYC2016TCSVT} takes $0.4s$, and GOG~\cite{Matsukawa2016CVPR} takes $1.3s$. As for the MSTC, the time for computing the scores between one probe and all gallery is $3.8\times10^{-3}$ seconds. While DNS~\cite{ZhangL2016CVPR}, CVDCA~\cite{ChenYC2016TCSVT}, and KCVDCA~\cite{ChenYC2016TCSVT} takes $5.4\times10^{-3}s$, $6.0\times10^{-3}s$, and $148.2\times10^{-3}s$, respectively, which confirms the efficiency of the proposed method. On the CUHK03 dataset, the testing time for MSTC is $2.5\times10^{-2}s$ per probe, while the MuDeep~\cite{Qian2017ICCV} takes $4.0\times10^{-2}s$. As for the number of parameters, the MSTC has 1.18 million parameters, while the MuDeep~\cite{Qian2017ICCV} has 138.02 million parameters.
\begin{table}[!t]
\renewcommand{\arraystretch}{1.2}
\centering
\caption{Top Recognition rate (in percentage) of the cross-dataset experiment.}
\begin{tabular}{|c|c|ccc|}
	\hline
Source/Target                                           &Methods                        & r=1                   & r=5               & r=10              \\ \hline
\multirow{3}*{\rotatebox{0}{VIPeR/Market-1501}}         &Ours                           & \textbf{24.13}        & \textbf{38.93}    & \textbf{46.85}             \\
~                                                       &MuDeep\cite{Qian2017ICCV}      & 18.70                 & 34.40             & 41.90             \\
~                                                       &HA-CNN\cite{Li2018CVPR}        & 18.10                 & 32.10             & 39.60             \\
\hline
\multirow{3}*{\rotatebox{0}{Market-1501/VIPeR}}         &Ours                           & \textbf{18.35}        & \textbf{32.59}    & \textbf{38.29}             \\
~                                                       &MuDeep\cite{Qian2017ICCV}      & 7.80                  & 13.90             & 19.00              \\
~                                                       &HA-CNN\cite{Li2018CVPR}        & 15.70                 & 25.00             & 31.20             \\
 \hline
\end{tabular}
\label{tab-crossdata}
\end{table}
\subsubsection{Challenge Cases}
Four challenging cases are shown in Fig.~\ref{img-challengecase}, including occlusion, multi-pedestrian, bad view and low resolution. For the occlusion problem, the semantic feature extracted from the visible region of body could provide a vital cue to re-identify the matched one. Besides, the global and local region features used by the SRR also improve the robustness to the partial occlusion. As to the multi-pedestrian problem, the semantic information is extracted from the primary pedestrian in the foreground, which avoids the interference of other pedestrian body parts in the image. Thus, the right one could obtain a satisfactory ranking in the cases of occlusion and multi-pedestrian, as shown in the first and second rows of Fig.~\ref{img-challengecase}. However, there also exists failure cases such as bad view and low resolution. As shown in the third row of Fig.~\ref{img-challengecase}, there exists a large appearance variance between the front and back perspective, which makes it difficult to match the right one. The low image quality also increases the challenge of person re-identification as shown in the last row of Fig~\ref{img-challengecase}. Although the serious interferences increase the challenge of the task, it can be observed that the matched pedestrian has a relatively well ranking with the help of the proposed method.
\subsubsection{Cross-dataset evaluation}
To investigate the generalization of the proposed MSTC, the cross-dataset experiment is conducted. The learned metric is trained and tested on different datasets. Since there is a large variation in the data distribution of different datasets, simply transferring the metric trained in source domain to target domain could not get a satisfactory result. Table~\ref{tab-crossdata} shows the performance comparisons of the proposed method and two deep learning baseline (i.e. MuDeep~\cite{Qian2017ICCV}, HA-CNN~\cite{Li2018CVPR}). As can be seen in Table~\ref{tab-crossdata}, the proposed method produces better results at different ranks.

\section{Conclusion}
In this paper, we propose a reliable feature SRR, together with a comprehensive constraint MSTC, for the person re-identification task. The semantic information is applied to find the corresponding parts between image pairs, which helps the SRR tackle the misalignment problem caused by viewpoint changes and pose variations. Besides, the comprehensive constraint MSTC is proposed to learn a more discriminative metric, which exploits the topological relationship among probe and gallery samples to maintain the compactness of intra-class as well as the sparsity of inter-class. Extensive experiments on five challenging datasets have shown that the proposed method achieves competitive performance with the state-of-the-art algorithms, which demonstrates the effectiveness of the proposed method.

In the future, there are several interesting directions along which we would extend the work. The first is to extend the proposed method to video based person re-identification task, which utilizes the spatial temporal cue to measure the similarity of pair sequences of pedestrian images. Another one is to further generalize the MSTC constraint to other applications with deep-learned features.

\ifCLASSOPTIONcaptionsoff
  \newpage
\fi


\begin{thebibliography}{99}

\bibitem{Garcia2017tip}
J. Garcia, N. Martinel, A. Gardel, I. Bravo, G. L. Foresti, \emph{et al.}, ``Discriminant Context Information Analysis for Post-Ranking Person Re-Identification,'' \emph{IEEE Trans. Image Process.}, vol. 26, no. 4, pp. 1650-1665, 2017.

\bibitem{Pang2016TIE}
Y. Pang, H. Zhu, X. Li, and J. Pan, ``Motion Blur Detection with an Indicator Function for Surveillance Robots,'' \emph{IEEE Transactions on Industrial Electronics.}, vol. 63, no. 9, pp. 5592-5601, 2016.

\bibitem{Wang2016TCSVT}
X. Wang, W.-S. Zheng, X. Li, and J. Zhang, ``Cross-Scenario Transfer Person Re-Identification,'' \emph{IEEE Trans. Circuits and Systems for Video Technology.}, vol. 26, no. 8, pp. 1447-1460, 2016.

\bibitem{Zhang2016arxiv}
L. Zhang, Y. Yang, and A. G. Hauptmann, ``Person Re-Identification: Past, Present and Future,'' \emph{arXiv preprint arXiv:} 1610.02984, 2016.

\bibitem{Chen2016TCSVT}
W. Chen, L. Cao, X. Chen, and K. Huang, ``An Equalized Global Graph Model-Based Approach for Multi-Camera Object Tracking,'' \emph{IEEE Trans. Circuits and Systems for Video Technology.}, vol. 27, no. 11, pp. 2367-2381, 2017.

\bibitem{An2017TCSVT}
L. An, Z. Qin,  X. Chen, and S.-F. Yang, ``Multi-Level Common Space Learning for Person Re-Identification,'' \emph{IEEE Trans. Circuits and Systems for Video Technology.}, doi: 10.1109/TCSVT.2017.2680118, pp. 1-11, 2017.

\bibitem{Zhu2017TCSVT}
J. Zhu, H. Zeng, S. Liao, Z. Lei, C. Cai, \emph{et al.}, ``Deep Hybrid Similarity Learning for Person Re-Identification,'' \emph{IEEE Trans. Circuits and Systems for Video Technology.}, doi: 10.1109/TCSVT.2017.2734740, pp. 1-11, 2017.

\bibitem{Zheng2016IJCV}
L. Zheng, S. Wang, J. Wang, and Q. Tian, ``Accurate Image Search with Multi-Scale Contextual Evidences,'' \emph{International Journal of Computer Vision}, vol. 120, no. 1, pp. 1-13, 2016.

\bibitem{Zhu2017TIP}
F. Zhu, X. Kong, L. Zheng, H. Fu, and Q. Tian, ``Part-Based Deep Hashing for Large-Scale Person Re-Identification,'' \emph{IEEE Trans. Image Process.}, vol. 26, no. 10, pp. 4806-4817, 2017.

\bibitem{Liu2015Neuro}
H. Liu, B. Ma, L. Qin, J. Pang, and Q. Huang, ``Set-Label Modeling and Deep Metric Learning on Person Re-Identification,'' \emph{Neurocomputing.}, vol. 151, pp. 1283-1292, 2015.

\bibitem{Tao2017TCSVT}
D. Tao, Y. Guo, B. Yu, J. Pang, and Z. Yu, ``Deep Multi-View Feature Learning for Person Re-Identification,''  \emph{IEEE Trans. Circuits and Systems for Video Technology.}, doi: 10.1109/TCSVT.2017.2726580, pp. 1-11, 2017.

\bibitem{Zheng2014TIP}
L. Zheng, S. Wang, and Q. Tian, ``Coupled Binary Embedding for Large-Scale Image Retrieval,'' \emph{IEEE Trans. Image Process.}, vol. 23, no. 8, pp. 3368-3380, 2014.

\bibitem{Jing2017TIP}
X.-Y. Jing, X. Zhu, F. Wu, R. Hu, X. You, \emph{et al.}, ``Super-Resolution Person Re-Identification with Semi-Coupled Low-Rank Discriminant Dictionary Learning,'' \emph{IEEE Trans. Image Process.}, vol. 26, no. 3, pp. 1363-1378, 2017.

\bibitem{Liu2013ICIP}
H. Liu, L. Qin, Z. Cheng, and Q. Huang, ``Set-Based Classification for Person Re-Identification Utilizing Mutual-Information,'' \emph{in Proc. IEEE Int. Conf. Image Process.}, 2013, pp. 3078-3082.

\bibitem{Zheng2018TPAMI}
L. Zheng, Y. Yang, and Q. Tian, ``SIFT Meets CNN A Decade Survey of Instance Retrieval,'' \emph{IEEE Trans. Patt. Anal. Mach. Intell.}, vol. 40, no. 5, pp. 1224-1244, 2018.

\bibitem{Yang2014ECCV}
Y. Yang, J. Yang, J. Yan, S. Liao, D. Yi, \emph{et al.}, ``Salient Color Names for Person Re-Identification,'' \emph{in Proc. Eur. Conf. Comput. Vis.}, 2014, pp. 536-551.

\bibitem{Prates2016ICIP}
R. Prates, C. R. S. Dutra, and W. R. Schwartz, ``Predominant Color Name Indexing Structure for Person Re-Identification,'' \emph{in Proc. IEEE Int. Conf. Image Process.}, 2016, pp. 779-783.

\bibitem{Matsukawa2016CVPR}
T. Matsukawa, T. Okabe, E. Suzuki, and Y. Sato, ``Hierarchical Gaussian Descriptor for Person Re-Identification,'' \emph{in Proc. IEEE Conf. Comput. Vis. Pattern Recog.}, 2016, pp. 1363-1372.

\bibitem{Lin2017TIP}
W. Lin, Y. Shen, J. Yan, M. Xu, J. Wu, \emph{et al.}, ``Learning Correspondence Structures for Person Re-Identification,'' \emph{IEEE Trans. Image Process.}, vol. 26, no. 5, pp. 2438-2453, 2017.

\bibitem{Liao2015CVPR}
S. Liao, Y. Hu, X. Zhu, and S. Li, ``Person Re-Identification by Local Maximal Occurrence Representation and Metric Learning,''\emph{ in Proc. IEEE Conf. Comput. Vis. Pattern Recog.}, 2015, pp. 2197-2206.

\bibitem{Ye2016TMM}
M. Ye, C. Liang, Y. Yu, Z. Wang, Q. Leng, \emph{et al.}, ``Person Re-Identification via Ranking Aggregation of Similarity Pulling and Dissimilarity Pushing,'' \emph{IEEE Trans. Multimedia.}, vol. 18, no. 12, pp. 2553-2566, 2016.

\bibitem{Chen2016CVPR}
D. Chen, Z. Yuan, B. Chen, and N. Zheng, ``Similarity Learning with Spatial Constraints for Person Re-Identification,'' \emph{in Proc. IEEE Conf. Comput. Vis. Pattern Recog.}, 2016, pp. 1268-1277.

\bibitem{Tao2016TIP}
D. Tao, Y. Guo, M. Song, Y. Li, Z. Yu, \emph{et al.}, ``Person Re-Identification by Dual-Regularized KISS Metric Learning,'' \emph{IEEE Trans. Image Process.}, vol. 25, no. 6, pp. 2726-2738, 2016.

\bibitem{Li2016TCSVT}
S.-M. Li, C. Gao, J.-G. Zhu, and C. Li, ``Person Re-Identification Using Attribute-Restricted Projection Metric Learning,'' \emph{IEEE Trans. Circuits and Systems for Video Technology.}, doi: 10.1109/TCSVT.2016.2637819, pp. 1-12, 2016.

\bibitem{Farenzena2010CVPR}
M. Farenzena, L. Bazzani, A. Perina, V. Murino, and M. Cristani, ``Person Re-Identification by Symmetry-Driven Accumulation of Local Features,'' \emph{in Proc. IEEE Conf. Comput. Vis. Pattern Recog.}, 2010, pp. 2360-2367.

\bibitem{Mignon2012CVPR}
A. Mignon and F. Jurie, ``PCCA: A New Approach for Distance Learning from Sparse Pairwise Constraints,'' \emph{in Proc. IEEE Conf. Comput. Vis. Pattern Recog.}, 2012, pp. 2666-2672.

\bibitem{Su2017TPAMI}
C. Su, F. Yang, S. Zhang, Q. Tian, L. S. Davis, \emph{et al.}, ``Multi-Task Learning with Low Rank Attribute Embedding for Multi-Camera Person Re-Identification,'' \emph{IEEE Trans. Patt. Anal. Mach. Intell.}, vol. 40, no. 5, pp. 1167-1181, 2018.

\bibitem{Zhao2017TPAMI}
R. Zhao, W. Oyang, and X. Wang, ``Person Re-Identification by Saliency Learning,'' \emph{IEEE Trans. Patt. Anal. Mach. Intell.}, vol. 39, no. 2, pp. 356-370, 2017.

\bibitem{Cheng2016CVPR}
D. Cheng, Y. Gong, S. Zhou, J. Wang, and N. Zheng, ``Person Re-Identification by Multi-Channel Parts-Based CNN with Improved Triplet Loss Function,'' \emph{in Proc. IEEE Conf. Comput. Vis. Pattern Recog.}, 2016, pp. 1335-1344.

\bibitem{Su2016ECCV}
C. Su, S. Zhang, J. Xing, W. Gao, and Q. Tian, ``Deep Attributes Driven Multi-Camera Person Re-Identification,'' \emph{in  Proc. Eur. Conf. Comput. Vis.}, 2016, pp. 475-491.

\bibitem{Wu2016WACV}
S. Wu, Y.-C. Chen, X. Li, A.-C. Wu, J.-J. You, and W.-S. Zheng, ``An Enhanced Deep Feature Representation for Person Re-Identification,'' \emph{in Proc. IEEE Winter Conference on Applications of computer vision.}, 2016, pp. 1-8.

\bibitem{LiD2017ICCV}
D. Li, X. Chen, Z. Zhang, and K. Huang, ``Learning Deep Context-aware Features over Body and Latent Parts for Person Re-identification,'' \emph{in Proc. IEEE Conf. Comput. Vis. Pattern Recog.}, 2017, pp. 7398-7407.

\bibitem{ZhaoH2017ICCV}
H. Zhao, M. Tian, S. Sun, J. Shao, J. Yan, \emph{et al}, ``Spindle Net: Person Re-identification with Human Body Region Guided Feature Decomposition and Fusion,'' \emph{in Proc. IEEE Conf. Comput. Vis. Pattern Recog.}, 2017, pp. 907-915.

\bibitem{Zhao2017ICCV}
L. Zhao, X. Li, J. Wang, and Y. Zhuang, ``Deeply-Learned Part-Aligned Representations for Person Re-Identification,'' \emph{in Proc. IEEE Int. Conf. Comput. Vis.}, 2017, pp. 3289-3248.

\bibitem{Su2017ICCV}
C. Su, J. Li, S. Zhang, J. Xing, and W. Gao, ``Pose-driven Deep Convolutional Model for Person Re-identification,'' \emph{in Proc. IEEE Int. Conf. Comput. Vis.}, 2017, pp. 3980-3989.

\bibitem{Bak2017TCSVT}
S. Bak and P. Carr, ``Deep Deformable Patch Metric Learning for Person Re-identification,'' \emph{IEEE Trans. Circuits and Systems for Video Technology}, doi: 10.1109/TCSVT.2017.2765242, pp.1-12, 2017.

\bibitem{Zheng2017Arxiv}
L. Zheng, Y. Huang, H. Lu, and Y. Yang, ``Pose Invariant Embedding for Deep Person Re-identification,'' \emph{arXiv preprint arXiv:} 1701.07732, 2017.

\bibitem{Sun2017Arxiv}
Y. Sun, L. Zheng, Y. Yang, Q. Tian, and S. Wang, ``Beyond Part Models: Person Retrieval with Refined Part Pooling (and a Strong Convolutional Baseline),'' \emph{arXiv preprint arXiv:} 1711.09349, 2017.

\bibitem{Wang2014TCSVT}
Y. Wang, R. Hu, C. Liang, C. Zhang, and Q. Leng, ``Camera Compensation Using a Feature Projection Matrix for Person Re-Identification,'' \emph{IEEE Trans. Circuits and Systems for Video Technology.}, vol. 24, no. 8, pp. 1350-1361, 2014.

\bibitem{Hirzer2012CVPR}
M. Hirzer, ``Large Scale Metric Learning from Equivalence Constraints,'' \emph{in Proc. IEEE Conf. Comput. Vis. Pattern Recog.}, 2012, pp. 2288-2295.

\bibitem{ZhangL2016CVPR}
L. Zhang, T. Xiang, and S. Gong, ``Learning a Discriminative Null Space for Person Re-Identification,'' \emph{in Proc. IEEE Conf. Comput. Vis. Pattern Recog.}, 2016, pp. 1239-1248.

\bibitem{Li2016ICIP}
Z. Li, Z. Han, and Q. Ye, ``Person Re-Identification via Adaboost Ranking Ensemble,'' \emph{in Proc. IEEE Int. Conf. Image Process.}, 2016, pp. 4269-4273.

\bibitem{Weinberger2005NIPS}
K. Q. Weinberger and L. K. Saul, ``Distance Metric Learning for Large Margin Nearest Neighbor Classification,'' \emph{in Advances in neural information processing systems.}, 2005, pp.1473-1480.

\bibitem{Martinel2015TPAMI}
N. Martinel, A. Das, C. Micheloni, and A. K. Roy-Chowdhury, ``Re-Identification in the Function Space of Feature Warps,'' \emph{IEEE Trans. Patt. Anal. Mach. Intell.}, vol. 37, no. 8, pp. 1656-1669, 2015.

\bibitem{Wang2016ICIP}
J. Wang, J. Zhu, Z. Wang, C. Gao, N. Sang, \emph{et al.}, ``Contextual Similarity Regularized Metric Learning for Person Re-Identification,'' \emph{in Proc. IEEE Int. Conf. Image Process.}, 2016, pp. 2048-2053.

\bibitem{Zhou2017CVPR}
S. Zhou, J. Wang, J. Wang, Y. Gong, and N. Zheng, ``Point to Set Similarity Based Deep Feature Learning for Person Re-identification,'' \emph{in Proc. IEEE Conf. Comput. Vis. Pattern Recog.}, 2017, pp. 5028-5037.

\bibitem{Chen2017ICCV}
J. Chen, Y. Wang, J. Qin, L. Liu, and L. Shao, ``Fast Person Re-identification via Cross-camera Semantic Binary Transformation,'' \emph{in Proc. IEEE Conf. Comput. Vis. Pattern Recog.}, 2017, pp. 5330-5339.

\bibitem{Bak2017CVPR}
S. Bak and P. Carr, ``One-Shot Metric Learning for Person Re-identification,'' \emph{in Proc. IEEE Conf. Comput. Vis. Pattern Recog.}, 2017, pp. 1571-1580.

\bibitem{Wen2016ECCV}
Y. Wen, K. Zhang, Z. Li, and Y. Qiao, ``A Discriminative Feature Learning Approach for Deep Face Recognition,'' \emph{in Proc. Eur. Conf. Comput. Vis.}, 2016, pp. 499-515.

\bibitem{Jin2017IJCB}
H. Jin, X. Wang, S. Liao, and S.-Z. Li, ``Deep person re-identification with improved embedding and efficient training,'' \emph{IEEE International Joint Conference on Biometrics.}, 2017, pp. 261-267.

\bibitem{Liu2017CVPR}
W. Liu, Y. Wen, Z. Yu, M. Li, B. Raj, \emph{et al}, ``SphereFace: Deep Hypersphere Embedding for Face Recognition,'' \emph{in Proc. IEEE Conf. Comput. Vis. Pattern Recog.}, 2017, pp. 6738-6746.

\bibitem{Zhong2017CVPR}
Z. Zhong, L. Zheng, D. Cao, and S. Li, ``Re-Ranking Person Re-Identification With k-Reciprocal Encoding,'' \emph{in Proc. IEEE Conf. Comput. Vis. Pattern Recog.}, 2017, pp. 3652-3661.

\bibitem{Hermans2017arxiv}
A. Hermans, L. Beyer, and B. Leibe, ``In Defense of the Triplet Loss for Person Re-Identification,'' \emph{arXiv preprint arXiv:} 1703.07737, 2017.

\bibitem{Yang2011CVPR}
Y. Yang and D. Ramanan, ``Articulated Pose Estimation with Flexible Mixtures-of-Parts,'' \emph{in Proc. IEEE Conf. Comput. Vis. Pattern Recog.}, 2011, pp. 1385-1392.

\bibitem{Liao2010CVPR}
S. Liao, G. Zhao, V. Kellokumpu, M. Pietikainen, and S. Li, ``Modeling Pixel Process with Scale Invariant Local Patterns for Background Subtraction in Complex Scenes,'' \emph{in Proc. IEEE Conf. Comput. Vis. Pattern Recog.}, 2010, pp. 1301-1306.

\bibitem{chen2015CVPR}
D. Chen, Z. Yuan, G. Hua, N. Zheng, and J. Wang, ``Similarity Learning on an Explicit Polynomial Kernel Feature Map for Person Re-Identification,'' \emph{in Proc. IEEE Conf. Comput. Vis. Pattern Recog.}, 2015, pp. 1565-1573.

\bibitem{Liu2013TPAMI}
G. Liu, Z. Lin, S. Yan, J. Sun, Y. Yu, \emph{et al.}, ``Robust Recovery of Subspace Structures by Low-Rank Representation,'' \emph{IEEE Trans. Patt. Anal. Mach. Intell.}, vol. 35, no. 1, pp. 171-184, 2013.

\bibitem{Boyd2011}
S. Boyd, N. Parikh, E. Chu, B. Peleato, and J. Eckstein, ``Distributed Optimization and Statistical Learning via the Alternating Direction Method of Multipliers,'' \emph{Foundations \& Trends in Machine Learning.}, vol. 3, no. 1, pp. 1-122, 2011.

\bibitem{Kowalski2009ACHA}
M. Kowalski, ``Sparse Regression Using Mixed Norms,'' \emph{Applied \& Computational Harmonic Analysis.}, vol. 27, no. 3, pp. 303-324, 2009.

\bibitem{Dattorro2006AMathS}
J. Dattorro, ``Convex Optimization \& Euclidean Distance Geometry,'' \emph{Applied Mathematical Sciences.}, vol. 51, no. 11, 2006.

\bibitem{Gray2008ECCV}
D. Gray and H. Tao, ``Viewpoint Invariant Pedestrian Recognition with an Ensemble of Localized Features,'' \emph{in Proc. Eur. Conf. Comput. Vis.}, 2008, pp. 262-275.
\bibitem{Guo2016ICPR}
C.-C. Guo, S.-Z.Chen, J.-H. Lai, X.-J. Hu, and S.-C. Shi, ``Multi-Shot Person Re-Identification with Automatic Ambiguity Inference and Removal,'' \emph{in Proc. IEEE Int. Conf. Pattern Recog.}, 2016, pp. 3540-3545.
\bibitem{Chen2009CVPR}
C.-L. Chen, X. Tao, and S. Gong, ``Multi-Camera Activity Correlation Analysis,'' \emph{in Proc. IEEE Conf. Comput. Vis. Pattern Recog.}, 2009, pp. 1988-1995.

\bibitem{Li2014CVPR}
W. Li, R. Zhao, T. Xiao, and X. Wang, ``DeepReID: Deep Filter Pairing Neural Network for Person Re-identification,'' \emph{in Proc. IEEE Conf. Comput. Vis. Pattern Recog.}, 2014, pp. 152-159.

\bibitem{Zheng2016ICCV}
L. Zheng, L. Shen, L. Tian, S. Wang, J. Wang, \emph{et al}, ``Scalable person re-identification: A benchmark,'' \emph{in Proc. IEEE Int. Conf. Comput. Vis.}, 2016, pp. 1116-1124.

\bibitem{Wang2007ICCV}
X. Wang, G. Doretto, T. Sebastian, J. Rittscher, and P. Tu, ``Shape and Appearance Context Modeling,'' \emph{in Proc. IEEE Int. Conf. Comput. Vis.}, 2007, pp. 1-8.

\bibitem{ZhangY2016CVPR}
Y. Zhang, B. Li, H. Lu, A. Irie, and R. Xiang, ``Sample-Specific SVM Learning for Person Re-Identification,'' \emph{in Proc. IEEE Conf. Comput. Vis. Pattern Recog.}, 2016, pp. 1278-1287.

\bibitem{ChenYC2016TCSVT}
Y.-C. Chen, W.-S. Zheng, J.-H. Lai, and P.-C. Yuen, ``An Asymmetric Distance Model for Cross-View Feature Mapping in Person Re-Identification,'' \emph{IEEE Trans. Circuits and Systems for Video Technology}, vol. 27, no. 8, pp. 1661-1675, 2016.

\bibitem{Xiong2014ECCV}
F. Xiong, M. Gou, O. Camps, and M. Sznaier, ``Person Re-Identification Using Kernel-Based Metric Learning Methods,'' \emph{in Eur. Conf. Comput. Vis.}, 2014, pp. 1-16.

\bibitem{Qian2017ICCV}
X. Qian, Y. Fu, Y.-G. Jiang,  T. Xiang, and X. Xue, ``Multi-Scale Deep Learning Architectures for Person Re-Identification,'' \emph{in Proc. IEEE Int. Conf. Comput. Vis.}, 2017, pp. 5409-5418.

\bibitem{Peng2018TPAMI}
P. Peng, Y. Tian, T. Xiang, Y. Wang, M. Pontil, \emph{et al.}, ``Joint Semantic and Latent Attribute Modelling for Cross-Class Transfer Learning,'' \emph{IEEE Trans. Patt. Anal. Mach. Intell.}, doi: 10.1109/TPAMI.2017.2723882, pp.1-14, 2018.

\bibitem{Zhou2017ICCV}
J. Zhou, P. Yu, W. Tang, and Y. Wu, ``Efficient Online Local Metric Adaptation via Negative Samples for Person Re-Identification,'' \emph{in Proc. IEEE Int. Conf. Comput. Vis.}, 2017, pp. 2439-2447.

\bibitem{Zhou2018TMM}
S. Zhou, J. Wang, R. Shi, Q. Hou, Y. Gong, \emph{et al.}, ``Large Margin Learning in Set to Set Similarity Comparison for Person Re-identification,'' \emph{IEEE Trans. Multimedia.}, vol. 20, no. 3, pp. 593-604, Sep, 2018.

\bibitem{Zhong2018CVPR}
Z. Zhong, L. Zheng, Z. Zheng, S. Li, and Y. Yang, ``Camera Style Adaptation for Person Re-identification,'' \emph{in Proc. IEEE Conf. Comput. Vis. Pattern Recog.}, 2018.
\bibitem{Li2018CVPR}
W. Li, X. Zhu, and S. Gong, ``Harmonious Attention Network for Person Re-Identification,'' \emph{in Proc. IEEE Conf. Comput. Vis. Pattern Recog.}, 2018.

\bibitem{Zheng2017TOMM}
Z. Zheng, L. Zheng, and Y. Yang, ``A Discriminatively Learned CNN Embedding for Person Re-identification,'' \emph{ACM Trans. Multimedia Comput. Commun. Appl.}, vol. 14, no. 1, pp. 1-20, 2017.

\bibitem{LiZ2013CVPR}
Z. Li, S. Chang, F. Liang, T.-S. Huang, L. Cao, \emph{et al}. ``Learning Locally-Adaptive Decision Functions for Person Verification,'' \emph{in Proc. IEEE Conf. Comput. Vis. Pattern Recog.}, 2013, pp. 3610-3617.


\end{thebibliography}
\end{document}